\documentclass[11pt,a4paper]{article}

\usepackage[utf8]{inputenc}
\usepackage[T1]{fontenc}
\usepackage{lmodern}
\usepackage[margin=2.5cm]{geometry}
\usepackage{amsmath}
\usepackage{graphicx}
\usepackage{booktabs}
\usepackage{array}
\usepackage{tabularx}
\usepackage{pifont}
\usepackage{authblk}
\usepackage[backend=biber,style=numeric-comp,sorting=none,giveninits=true,maxbibnames=99,texencoding=ascii]{biblatex}
\usepackage{url}
\usepackage[hidelinks]{hyperref}

\graphicspath{{./}}
\title{Standing on Open Shoulders: Regional Fine-Tuning of Open-Weight Camera-Trap Models for UK Wildlife with Modest Data and Compute}
\author[ ]{Paul Fergus}
\affil[ ]{Liverpool John Moores University, Byrom Street, Liverpool L3 3AF, United Kingdom}
\affil[ ]{Correspondence: \href{mailto:p.fergus@ljmu.ac.uk}{p.fergus@ljmu.ac.uk}}

\date{}

\begin{document}
\maketitle

\begin{abstract}
Camera-trap species classification is dominated by large models from well-resourced organisations, which often lack the species-level distinctions that regional conservation needs, such as distinguishing native red from invasive grey squirrels in the United Kingdom. We ask whether open weights allow a small team, with a modest dataset and a single graphics processing unit (GPU), to take a meaningful part in AI development. We fine-tuned the open DeepFaune classifier on 28 UK classes, paired it with the unmodified MegaDetector v6, and compared the resulting DeepFaune-UK v1.2 with DeepFaune v1.5.0 and SpeciesNet on hand-verified citizen-science images from the MammalWeb project. Because 1,039 of the 21,640 candidate images (4.8\%) also occurred in our training archive, every sequence containing one was removed, leaving 20,212 images. DeepFaune-UK achieved a strict image-level accuracy of 88.8\%, including correct identification of empty frames, against 56.5\% and 59.0\% for the baselines, but at DeepFaune's coarser label resolution DeepFaune v1.5.0 was more accurate on this set, though not significantly when contributors were resampled (91.4\% vs. 88.8\%). As DeepFaune v1.5.0 is built on a newer, DINOv3-based classifier, we tested whether this gap reflected fine-tuning or the starting model by applying the identical recipe, without new data, to that classifier. The resulting DeepFaune-UK v1.3 reached a strict accuracy of 90.6\%, narrowed the like-for-like gap to 0.8 points and reduced roe deer misclassified as fallow deer by 69\%. On a stricter subset of 1,850 images from the 26 contributors with no overlapping images, the three DeepFaune-based models did not differ significantly (90.5\%, 90.6\% and 89.9\% like-for-like). Fine-tuning on only 10\% of the training crops (4,270, about 150 per class on average) gave out-of-distribution accuracy similar to that obtained with all of them, and such a DINOv2 run took about 22 minutes with under 4 GB of GPU memory allocated. Open weights therefore make regional model development feasible for small teams, but the largest like-for-like gain on the main evaluation set among the factors we varied came from a newer base model trained by others, so this democratisation depends on large actors continuing to release open weights, under licence terms that can change.

\end{abstract}

\section{Introduction}

Britain's terrestrial mammals and ground-nesting birds continue to decline. The first comprehensive review of the status of British mammal populations in more than 20 years assessed all 58 of Britain's terrestrial mammal species~\parencite{mathews2018}, and the State of Nature 2023 assessment reaches a stark conclusion across taxa: one in six species is now threatened with extinction from Great Britain~\parencite{burns2023}. Camera traps have become the standard non-invasive tool for monitoring such populations, but the resulting image volume creates a review bottleneck: manual inspection is labour-intensive, delays conservation decisions, and does not scale to large or long-running deployments.

Deep learning has largely closed this gap. General-purpose animal detectors such as MegaDetector~\parencite{beery2019,hernandez2024} and large species classifiers such as DeepFaune~\parencite{rigoudy2023} and SpeciesNet~\parencite{gadot2024} now process millions of images for conservation organisations worldwide. These models are, however, the product of resources that few groups possess. SpeciesNet was trained on more than 65 million images~\parencite{google2025}; DeepFaune draws on images contributed by a consortium of more than 50 partners in France~\parencite{rigoudy2023}. Building models at this scale is beyond most universities, non-governmental organisations (NGOs) and community projects, reflecting a broader ``compute divide'' in artificial intelligence research~\parencite{ahmed2020}. At the same time, models trained for global or continental use inevitably reflect the priorities and data of their builders: species that matter locally may be absent, merged into coarse groups, or poorly represented, and an evaluation of camera-trap AI platforms on their users' own data, to which the present author contributed, found generally low to moderate species-level recall~\parencite{velez2023}.

The resolution at which a model names animals is not a cosmetic matter, because many wildlife management decisions in the UK turn on distinctions that coarse labels cannot make. The clearest example is the squirrel. The native red squirrel (\emph{Sciurus vulgaris}) has been displaced across much of Britain by the introduced North American grey squirrel (\emph{S. carolinensis}), which both outcompetes it and carries squirrelpox virus, a disease that rarely harms grey squirrels but is usually fatal to reds; red squirrel declines have been 17--25 times faster in regions where the virus is present in grey squirrel populations than in regions where it is not~\parencite{rushton2006}. The two species are accordingly treated in opposite ways in law: the red squirrel is protected under Schedule 5 of the Wildlife and Countryside Act 1981~\parencite{wca1981}, whereas the grey squirrel is a listed invasive alien species whose release into the wild is an offence in England and Wales~\parencite{ias2019}. Interventions aimed at one must therefore avoid the other. Oral fertility control, under development as a non-lethal means of reducing grey squirrel numbers, delivers bait to grey squirrels through feeders~\parencite{beatham2023}, and where both species occur the feeder must admit greys but exclude reds: the simple weighted door suitable for grey-only areas can be opened by larger red squirrels, so the design proposed for mixed areas uses an electronic weighing platform that opens the door only for animals above a minimum body weight, with a threshold of 450 g expected to exclude all red squirrels while admitting over 90\% of grey squirrels~\parencite{uksa2023}. A reliable visual classifier could complement or replace such mechanical discrimination, for example in an Internet-of-Things feeder that opens only when a grey squirrel is recognised and stays closed to reds, or a trap that is set only for greys. Species-selective devices of this kind already exist elsewhere: the Felixer grooming trap distinguishes feral cats and foxes from non-target wildlife before acting~\parencite{read2019}, and camera-equipped predator traps in New Zealand arm only when their AI identifies a target species with sufficient confidence~\parencite{pfnz2025}.

The need for species-level labels extends well beyond squirrels. The recovery of the native pine marten (\emph{Martes martes}) has been linked to the decline of grey squirrel populations, including a population crash in Ireland, and the recovery of reds in Ireland and Scotland~\parencite{sheehy2014,sheehy2018}, so monitoring martens and both squirrel species together indicates whether this natural form of control is under way. Among deer, the non-native Reeves' muntjac (\emph{Muntiacus reevesi}) is also a listed invasive species~\parencite{ias2019}, and in England and Wales the statutory close seasons that govern culling differ between the sexes and, for males, between roe deer and fallow, red and sika deer, while none applies to muntjac~\parencite{deeract1991}, so a survey that reports only ``deer'' cannot inform a lawful cull plan. For declining ground-nesting birds such as curlew (\emph{Numenius arquata}), a meta-analysis of nest-camera studies from Great Britain and the Isle of Man found that badger, red fox and, particularly on islands, hedgehog together accounted for 53\% of predation events on wader clutches~\parencite{barton2026}, and predator management depends on knowing which predator is present rather than merely that a mammal was recorded. Similar distinctions arise for species outside the scope of this study: between Scottish wildcats and their domestic and hybrid relatives, which are distinguished by diagnostic pelage characters~\parencite{kitchener2005}, and between the invasive American mink and native riparian mammals, mink presence having been found to be the main factor determining water vole distribution in the Thames catchment~\parencite{barreto1998}. In each case, the label a model returns determines what action can follow. Yet these are precisely the distinctions that large general-purpose models often do not make: DeepFaune reports any squirrel simply as ``squirrel'' and has no class for capercaillie (\emph{Tetrao urogallus}) or curlew, while SpeciesNet may fall back to a genus, family or broader label when uncertain. Regional adaptation offers a way to improve how consistently models return the species-level distinctions required by local monitoring.

The increasing release of open model weights offers a potential route around this impasse. If a small team can take openly released detectors and classifiers and adapt them to regional needs using a modest dataset and a single graphics processing unit (GPU), then participation in AI development need not be restricted to the best-resourced institutions. This is one specific sense in which AI may be ``democratised'' --- the democratisation of AI \emph{development}, as distinct from the democratisation of AI use, profits or governance~\parencite{seger2023}. Whether this promise holds in practice, and what its limits are, is an empirical question.

This paper is part of an ongoing programme applying AI-driven camera-trap monitoring to UK wildlife, which has previously delivered a real-time curlew monitoring system with the Game \& Wildlife Conservation Trust~\parencite{chalmers2025} and a single-stage detector for UK mammals and birds, released for non-commercial use and intended to make camera-trap AI accessible to non-specialist ecologists~\parencite{fergus2026}, deployed through the self-hostable Trap Tracker platform~\parencite{traptracker2026,traptrackeroffline2026}. Here we examine the democratisation question from the developer's side, through a concrete case study. We build DeepFaune-UK v1.2, a two-stage pipeline that combines the openly released MegaDetector v6 detector, used without modification, with the open DeepFaune vision-transformer classifier fine-tuned on 28 UK classes using 42,719 training crops on a single GPU. We compare it with the latest DeepFaune release (v1.5.0) and with SpeciesNet on hand-verified UK citizen-science images from the MammalWeb project~\parencite{hsing2018}, from which every sequence containing an image that also occurs in our training archive was removed (Section 3.5). Because the latest DeepFaune release also moves to a newer foundation backbone, we additionally apply the identical fine-tuning recipe to its DINOv3-based classifier, producing DeepFaune-UK v1.3, so that the contribution of the starting checkpoint can be separated from that of regional fine-tuning.

Two research questions guide the study. The first concerns performance: how does a regionally fine-tuned open-weight model compare with large, globally or continentally trained models on out-of-distribution UK camera-trap images, both at the level of individual species and on a like-for-like basis that accounts for differences in label vocabulary (RQ1)? The second concerns resources: how much data and compute does regional fine-tuning actually require, and how does out-of-distribution performance change as the training set grows (RQ2)? RQ1 is the question a practitioner would ask before adopting a model, but it is RQ2 on which the democratisation argument turns, since a regional model that matched its continental counterpart only after years of labelling and substantial compute would offer little encouragement to small teams.

In addressing these questions, the paper makes three contributions. First, it provides an out-of-distribution benchmark of open camera-trap models (two DeepFaune-UK versions, DeepFaune v1.5.0 and SpeciesNet) on hand-verified UK data, using a like-for-like protocol that accounts for differences in label vocabulary, removing evaluation sequences that overlap with the training archive and reporting a site-disjoint subset, and reporting confidence intervals and paired tests that resample whole camera-trap sequences or contributors, so that differences between models can be judged against their uncertainty. Second, it presents a learning-curve and resource-footprint analysis of regional fine-tuning on a single GPU, repeated over three random seeds, in which training time, peak GPU memory and measured GPU energy are reported alongside accuracy, together with a backbone comparison that repeats the experiment with the newer DINOv3-based classifier. Third, it offers a critical discussion of what these results do and do not imply for the democratisation of AI development in conservation, arguing that the access they demonstrate is real but conditional on the continued release of open weights and on the licences under which those weights are shared.

The remainder of this paper is organised as follows. Section 2 reviews automated camera-trap analysis, foundation models and open weights, and the democratisation debate. Section 3 describes the data, the DeepFaune-UK pipelines, the baselines, the out-of-distribution set, the evaluation protocol and the learning-curve experiment. Section 4 reports results; Section 5 discusses their implications and limitations; Section 6 concludes.

\section{Related Work}

Early work demonstrated that convolutional neural networks could identify species in camera-trap images at near-human accuracy when trained on large labelled collections such as Snapshot Serengeti~\parencite{norouzzadeh2018,willi2019}. A persistent difficulty is generalisation to new locations. Beery, Van Horn and Perona showed that camera-trap classifiers evaluated on held-out images from the same camera locations as training substantially over-report the accuracy achieved at genuinely new locations, largely because a fixed camera repeatedly photographs the same background, which models learn to exploit~\parencite{beery2018}. The evaluation of camera-trap AI platforms on users' own data cited above likewise found generally low to moderate species-level recall, concluded that most users would still need to review the predictions, and provided code for users to evaluate model performance on their own data~\parencite{velez2023}; the out-of-distribution evaluation in this paper follows the same principle.

This sensitivity to site has shaped the architecture of current systems. MegaDetector~\parencite{beery2019} established what is now the standard first stage for automated camera-trap review: a single, taxon-agnostic detector that localises animals and discards empty frames (later versions also detect people and vehicles), decoupling the comparatively easy and broadly transferable localisation problem from the harder, region-specific species identification problem. Its sixth version is designed with an emphasis on efficiency and is distributed through the PyTorch-Wildlife framework~\parencite{hernandez2024}. Two-stage pipelines, in which this detector is followed by a species classifier applied to each crop, are now widely used, and classifying detector crops rather than whole images improved macro-averaged F1 by around 25\% in a large comparison~\parencite{gadot2024}. SpeciesNet~\parencite{gadot2024,google2025} is a global classifier covering more than 2,000 labels, combined with MegaDetector and geographic filtering, which returns predictions at higher taxonomic ranks when uncertain.

The classifiers in such pipelines increasingly rest on a further layer of shared infrastructure. Self-supervised vision transformer models such as DINOv2~\parencite{oquab2024} and its successor DINOv3~\parencite{simeoni2025} learn general-purpose visual features from very large curated but unlabelled image collections, and transfer strongly to downstream classification tasks with frozen features or light fine-tuning. When such weights are released openly, downstream users can fine-tune them for new label sets at a small fraction of the original training cost. It is this property that makes regional adaptation plausible for groups that could never train such a backbone themselves. DeepFaune~\parencite{rigoudy2023} is the clearest example of this pattern in camera-trap ecology. Released with pretrained checkpoints by a consortium of over fifty French wildlife research, conservation and management partners, it is a two-stage recipe for European imagery in which a YOLO-family detector localises animal, person and vehicle boxes and a vision-transformer classifier, built on DINOv2 since release 1.1, classifies each crop into a European species vocabulary~\parencite{deepfaune2026}. Since release 1.4 it has offered an optional head that assigns birds to coarse groups, such as columbiform, galliform, corvid and raptor, and its latest release (v1.5.0) moves to a DINOv3 backbone~\parencite{deepfaune2026}. Because the checkpoints are openly released, they can serve as the starting point for a regional adaptation, which is the approach this paper adopts. Openness of this kind nevertheless comes with terms: the Creative Commons Attribution-ShareAlike licence under which the DeepFaune weights are currently distributed~\parencite{deepfaune2026}, the AGPL-3.0 licence of MegaDetector v6 and, for DeepFaune v1.5.0, the custom licence under which Meta distributes the DINOv3 backbone~\parencite{dinov3license} determine how derived models may be shared and deployed, a point to which we return in Section 5.

How much labelled data such adaptation requires is a practical question for any small team, and earlier studies gave cautious answers. Schneider et al. fine-tuned six ImageNet-pretrained convolutional networks on 47,279 Parks Canada images of 55 classes and found that classes with fewer than 500 training images had low and highly variable recall (0.750 $\pm$ 0.329), against 0.971 $\pm$ 0.014 for classes with more than 1,000, while the best model's top-1 accuracy fell from 95.6\% at the training locations to 68.7\% at unseen ones~\parencite{schneider2020}. Shahinfar et al. trained models with 10 to 1,000 images per class on datasets from Australia, Africa and North America and provided regression equations with which ecologists can estimate, before collecting data, how many images per species a target accuracy requires~\parencite{shahinfar2020}. Other work has reduced the labelling burden rather than the data requirement: MLWIC2 packages models trained on 3 million images in an R package whose Shiny applications let ecologists with minimal programming experience apply them and train new models on their own labelled images~\parencite{tabak2020}, and active learning matched state-of-the-art accuracy on a 3.2-million-image dataset with as few as 14,100 manual labels~\parencite{norouzzadeh2021}. These studies all used convolutional networks and predate self-supervised foundation backbones such as DINOv2. Whether such backbones change the amount of regional data needed, and whether the answer holds on independently labelled images from other sites, is the question addressed by the learning curve in this paper.

A different route to regional labels avoids fine-tuning altogether. Vision--language models pair images with text and can, in principle, recognise any species that can be named. BioCLIP, trained on the TreeOfLife-10M collection, substantially outperformed general-purpose baselines on fine-grained biological classification~\parencite{stevens2024}, and WildCLIP, obtained by fine-tuning CLIP on camera-trap images, retrieved such images by ecological attributes described in natural language~\parencite{gabeff2024}, but did so only after fine-tuning. Zero-shot recognition in camera-trap images has nevertheless proved fragile. An evaluation of BioCLIP on camera-trap wildlife images found that it struggled when applied without adaptation~\parencite{sajun2025}, and a study of 546 camera traps found that biological foundation models such as BioCLIP 2 underperformed at numerous sites, concluding that site-specific adaptation remains necessary~\parencite{jeon2026}. For a regional team that needs dependable species labels, particularly where a label triggers an action such as selective trapping, supervised fine-tuning of an open classifier therefore remains the more reliable route, and it is the one evaluated here.

Whether access to open weights amounts to democratisation depends on what the term is taken to mean. The phrase ``democratising AI'' is used with several distinct meanings. Seger et al. distinguish the democratisation of AI use, development, profits and governance, and caution against conflating democratisation with mere accessibility~\parencite{seger2023}. Ahmed and Wahed document a growing concentration of AI research in large firms and elite universities, which they attribute to unequal access to computing power~\parencite{ahmed2020}. Concerns about the financial and environmental costs of training large models~\parencite{strubell2019,schwartz2020} further motivate approaches that achieve useful results with modest resources and that report computational budgets alongside accuracy. A related literature questions what ``open'' means for AI models. Widder, Whittaker and West argue that claims of openness often elide the substantial concentration of industry power in large-scale AI development, and that openness alone, while it affords transparency, reuse and extension of existing models, does not perturb that concentration~\parencite{widder2024}. Liesenfeld and Dingemanse show that many models described as open are ``open weight'' at best, with information on training and fine-tuning data withheld, and argue that openness is composite and graded rather than a single property that a licence or download can establish~\parencite{liesenfeld2024}. Kapoor et al. identify customisability as a distinctive property of open foundation models, one that underpins their benefits for innovation and for the distribution of decision-making power~\parencite{kapoor2024}. Together, these accounts suggest that open weights can widen participation in adapting models while leaving their production concentrated, a distinction that this study examines empirically. In ecology, Tuia et al. argue that machine learning can transform wildlife conservation, but that this requires close collaboration between domain experts and data scientists~\parencite{tuia2022}. Our previous work addressed the democratisation of camera-trap AI \emph{use} by releasing a UK mammal and bird detector for non-specialist ecologists~\parencite{fergus2026}; the present study asks the complementary question of whether open foundation-model weights allow small, domain-led teams to participate in AI \emph{development}.

The question is not only an abstract one, because models built in this way can reach practitioners directly through open conservation infrastructure. The Conservation AI platform originated as a live-stream analysis system for anti-poaching drone video~\parencite{chalmers2019}, and was subsequently adapted to a real-time camera-trap pipeline, in which 3G/4G-enabled cameras send images to the platform as they are captured, deployed with the Game \& Wildlife Conservation Trust across 11 curlew nesting sites in Wales to support timely conservation intervention~\parencite{chalmers2025}. The same automated image-triage approach was adopted by the National Hedgehog Monitoring Programme, led by the British Hedgehog Preservation Society and the People's Trust for Endangered Species, to remove images without wildlife before manual review~\parencite{ljmu2024}, and related work has explored deriving richer ecological insights from camera-trap detections~\parencite{fergus2024}. Trap Tracker, the self-hostable successor to this infrastructure, provides an offline processor for batches of images~\parencite{traptrackeroffline2026} and a real-time application that analyses each new image as it arrives and raises alerts for watch-listed species~\parencite{traptracker2026}, both running user-supplied ONNX models. Models are not bundled with the software: users download them and place them in the application's model directory. DeepFaune-UK v1.2 and v1.3 are both released in this ONNX format~\parencite{fergus2026v12,fergus2026v13}, so the models evaluated here are exactly those that practitioners can download and deploy.

\section{Materials and Methods}

This section describes the training data and splitting procedure (3.1--3.2), the DeepFaune-UK v1.2 and v1.3 pipelines (3.3), the baseline models (3.4), the out-of-distribution evaluation set and its overlap check (3.5), the evaluation protocol (3.6), and the learning-curve and resource-footprint experiment (3.7). All training and inference ran inside Docker containers on a single NVIDIA RTX A6000 (48 GB) workstation GPU. Figure 1 summarises the development pipeline and the evaluation design.

\begin{figure}[htbp]
\centering
\includegraphics[width=\textwidth]{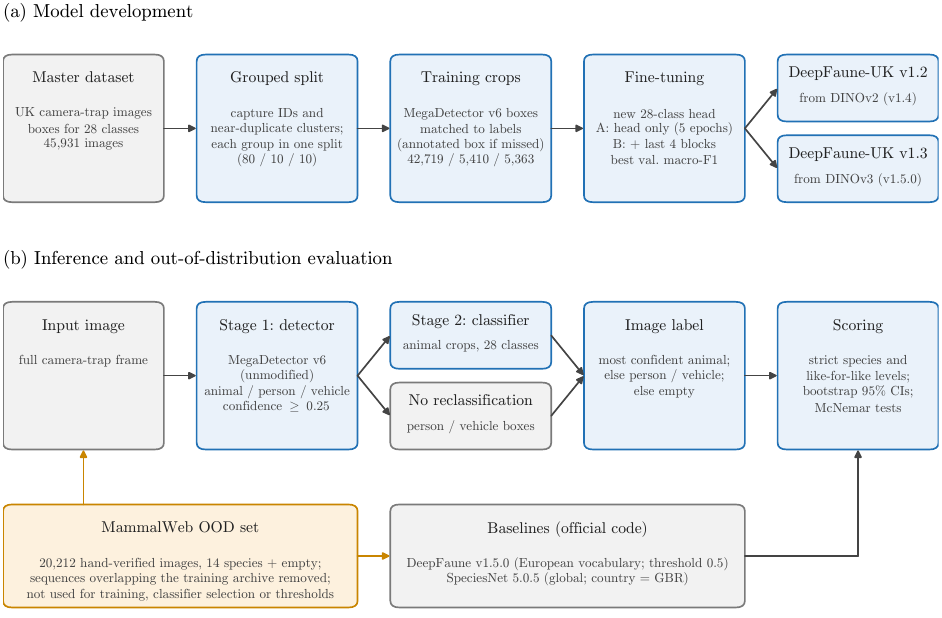}
\caption{Overview of the study. (a) Model development: the master dataset is split by capture group, MegaDetector v6 detections matched to the annotations (with the annotated box for any animal the detector missed) provide the training crops, and the DeepFaune classifier is fine-tuned in two phases from either its DINOv2 (v1.4) or its DINOv3 (v1.5.0) checkpoint. (b) Inference and evaluation: MegaDetector v6 localises animals, people and vehicles; only animal crops are classified, and the image label is that of the most confident animal detection. DeepFaune-UK and the two baselines are scored on the MammalWeb out-of-distribution set, which plays no part in training, classifier selection or threshold tuning (Sections 3.3 and 3.5).}
\label{fig:1}
\end{figure}

\subsection{Master Dataset}

The master dataset comprises UK camera-trap images with YOLO-format bounding-box annotations, assembled over a decade of Conservation AI and Trap Tracker deployments and the collaborations described in Section 2~\parencite{fergus2026}. Each image and its label file were renamed to the SHA-256 hash of the image's byte content, so that exact duplicates collapse automatically and the operation can be safely re-run as the archive grows. For DeepFaune-UK v1.2, the vocabulary was restricted to animal classes only: people, vehicles and the calibration poles used during camera deployment were removed, because people and vehicles are labelled directly by the detector in the deployed system (Section 3.3). The resulting vocabulary has 28 classes: 18 mammal species and 8 bird species, with capercaillie split into cock and hen, and curlew into adult and chick, because these distinctions matter for the monitoring programmes the model serves. After this restriction, 45,931 images remain.

\subsection{Leakage-Controlled Data Splitting}

An image-level random split is not a safe default for camera-trap data: a single animal encounter typically triggers a burst of near-identical frames seconds apart, and a fixed camera repeatedly photographs the same background for months or years~\parencite{beery2018}. The same phenomenon, near-duplicate or highly correlated samples on both sides of a split, inflates reported accuracy in image benchmarks generally~\parencite{barz2020}, accuracy on a benchmark's original test set can overstate accuracy on newly collected data~\parencite{recht2019}, and leakage of this kind is a recurring cause of overstated claims across applied machine learning~\parencite{kapoor2023}.

Images were therefore partitioned into groups, each assigned whole to a single split, using two complementary similarity signals combined by union-find clustering. First, capture identifiers were recovered from each image's original filename: a sequence-event hash embedded by some contributing sites' capture software, a Reconyx card identifier plus the frame counter grouped into blocks of ten consecutive values, or a site/camera token combined with a 10-minute timestamp bucket. This signal is precise but incomplete, covering 18,556 of the 48,364 images (38\%) in the archive before the restriction to animal classes; the grouping, allocation and leakage audit described here were run on this full archive. Second, to catch near-duplicates lacking usable metadata, every image pair was compared using the cosine similarity of whole-image embeddings from the DINOv2 ViT-L/14 backbone of the DeepFaune classifier (threshold $\geq$ 0.92) and a 64-bit perceptual hash (dHash, Hamming distance $\leq$ 2); a dHash match was accepted only when corroborated by embedding cosine similarity $\geq$ 0.80, because low-information night and over-exposed frames otherwise chain visually unrelated images together. Formally, two images $i$ and $j$ were linked if they shared a capture identifier or if
\begin{equation}
\cos(\mathbf{e}_i, \mathbf{e}_j) \geq 0.92 \quad \lor \quad \bigl[\, d_{\mathrm{H}}(h_i, h_j) \leq 2 \;\land\; \cos(\mathbf{e}_i, \mathbf{e}_j) \geq 0.80 \,\bigr],
\label{eq:links}
\end{equation}
where $\mathbf{e}_i$ is the L2-normalised embedding of image $i$, $h_i$ its 64-bit dHash and $d_{\mathrm{H}}$ the Hamming distance; the groups are the connected components of the resulting graph.

This grouping yielded 27,357 groups (24,321 singletons). Groups were assigned to training, validation and test sets by a greedy, species-stratified allocation targeting an 80/10/10 ratio. Species were processed in ascending order of their total image count, so that the rarest species claimed held-out groups before the validation and test quotas could be filled by common species. Within each species, groups were considered from smallest to largest and placed first in validation and then in test until that species reached 10\% of its images in each split; taking small groups first meets each quota closely and leaves the largest groups for training. A global cap of 15\% above the overall validation and test targets stopped a single large group containing several species from distorting the overall ratio, and all groups not placed by this procedure were assigned to training. The split was built incrementally as images were added to the archive: groups whose images already had a split kept it (26,864 groups in the final run), only new groups (493) were placed by this procedure, and any group bridging two splits would have been moved to training (none occurred). A post-hoc audit found no validation or test image whose dHash-nearest training image was within Hamming distance 2 with cosine $\geq$ 0.85, and none with embedding cosine $\geq$ 0.92 to any training image; a residual 16--18\% shared a camera framing with a training image while showing different content (the same fixed camera capturing a different animal at a different time). After restriction to the 28 animal classes, the splits contain 36,657 training, 4,626 validation and 4,648 test images. The split is frozen and reused unchanged by every experiment in this paper.

\subsection{DeepFaune-UK v1.2 and v1.3}

\textbf{Stage 1: detector.} Stage 1 is MegaDetector v6 (YOLOv10-e variant, 1280-pixel input)~\parencite{hernandez2024,pytorchwildlife2025}, used exactly as released, without fine-tuning. It was adopted in place of a fine-tuned DeepFaune YOLOv8s detector because it found substantially more animals in the out-of-distribution set described in Section 3.5: approximately 97\% vs. 84\% of its annotated animals (all 21,640 images). This detector comparison is the only use of that set that shaped the final pipeline, but it means that the set is not entirely independent of the pipeline's design (Section 5.1). The same set had earlier been used to compare superseded candidate systems (single-stage YOLOv10x and YOLO26x detectors, DeepFaune-UK v1.1 and a 30-class two-stage pipeline); these comparisons did not determine the final design. MegaDetector v6 assigns each box to one of three classes (animal, person or vehicle). Only animal boxes are passed to the Stage 2 classifier; person and vehicle boxes are reported with the detector's label and are not reclassified, which is why the classifier vocabulary contains animal classes only (Section 3.1).

\textbf{Stage-2 crop generation.} The classifier is trained on crops of detector output rather than ground-truth boxes, so that it sees inputs matching its deployment-time distribution. MegaDetector was run over every master image (confidence $\geq$ 0.25); each predicted box was greedily matched, highest confidence first, to an unused ground-truth box at Intersection-over-Union (IoU) $\geq$ 0.5, inheriting its species label while the crop was taken from the predicted box. Any ground-truth box the detector failed to match was cropped from its own coordinates, so that every annotated instance contributes a training crop; unmatched detector predictions were discarded. Each box was padded on every side by 20\% of its longer side, expanded to a square about its centre, clipped to the image border, and saved as JPEG with the longer side downscaled to at most 640 pixels. This produced 53,492 crops (42,719 training, 5,410 validation and 5,363 test): 51,207 from matched detector predictions and 2,285 from the ground-truth fallback.

\textbf{Stage 2: classifier.} Stage 2 is the DeepFaune ViT-L/14 classifier with a DINOv2 backbone (release v1.4)~\parencite{deepfaune2026,oquab2024}, whose original European-fauna classification head was discarded and replaced by a randomly initialised 28-way linear head. Input resolution (182 pixels) and normalisation statistics are read from the transform stored inside the DeepFaune checkpoint, so pre-processing matches what the pretrained backbone expects. Fine-tuning proceeds in two phases, following the principle that updating a large pretrained network too aggressively on a small downstream dataset risks overfitting~\parencite{yosinski2014} and catastrophic forgetting of the general features that motivated its use~\parencite{howard2018}. Phase A trains only the new head with the backbone frozen (AdamW~\parencite{loshchilov2019}, weight decay 0.05, constant learning rate 1 $\times$ 10$^{-3}$, 5 epochs). Phase B also unfreezes the last 4 of 24 transformer blocks and the final layer norm, at a learning rate of 1 $\times$ 10$^{-5}$ for up to 40 epochs, with early stopping (patience 10) on validation macro-F1. Both phases use a class-inverse-frequency weighted sampler to counteract class imbalance~\parencite{buda2018}, cross-entropy loss with label smoothing of 0.1~\parencite{szegedy2016,muller2019}, light augmentation (random resized crop, horizontal flip, mild colour jitter), mixed precision and a batch size of 64. The checkpoint with the highest validation macro-F1 is retained; macro-F1 rather than accuracy is used because it weights every species equally regardless of frequency.

\textbf{Cascade rule.} At inference, each animal box is cropped exactly as in training and classified; the classifier is restricted to animal species, and person and vehicle boxes retain the detector's label. Crop and normalisation parameters are stored in the exported ONNX model, so that deployment in Trap Tracker applies the same crop geometry, input size and normalisation as training. The production classifier trained for 23 epochs in 48 minutes (including the final test evaluation) and reached macro-F1 0.991 on the held-out test crops. The released model is available on Zenodo~\parencite{fergus2026v12}.

\textbf{DeepFaune-UK v1.3.} DeepFaune-UK v1.3 differs from v1.2 only in the starting point of the classifier. It is warm-started from the DeepFaune v1.5.0 checkpoint, a ViT-L/16 with a DINOv3 backbone~\parencite{simeoni2025} that classifies from the class token, at the 224-pixel input resolution and normalisation stored in that checkpoint; the European head is again replaced by a randomly initialised 28-way head. The detector, training crops, data splits, two-phase recipe and hyperparameters are identical to those of v1.2. Three full-data models were trained with different random seeds as part of the backbone comparison (Section 3.7), and the one with the highest validation macro-F1 was released as v1.3; the OOD set played no part in this choice. The selected model reached its best validation macro-F1 (0.994) after the 5 epochs of Phase A and 19 epochs of Phase B, trained in 107 minutes in total, and achieved test-crop macro-F1 0.994. It is released on Zenodo in the same ONNX format as v1.2~\parencite{fergus2026v13}.

\subsection{Baseline Models}

\textbf{DeepFaune v1.5.0} (released September 2026) was run with its official code~\parencite{deepfaune2026}: the DeepFaune YOLOv8s detector with MegaDetector v1000-sorrel as a backstop for images in which it finds no detection, the DINOv3 ViT-L classifier~\parencite{simeoni2025} (224-pixel input), the optional bird head (eight coarse bird groups) enabled, and a classification threshold of 0.5, as in the release's example prediction script (the graphical interface defaults to 0.8), below which images are labelled ``undefined''. Sequence-level smoothing was disabled so that all models were assessed on individual images.

\textbf{SpeciesNet} (package 5.0.5, model 4.0.3a)~\parencite{gadot2024,google2025} was run with its default ensemble of MegaDetector v5a and the SpeciesNet crop classifier, with geographic filtering (enabled by default) set to the United Kingdom (country code GBR).

Both baselines produce a single label per image. All models were evaluated at fixed, documented operating points without tuning on the OOD set: the defaults for SpeciesNet and DeepFaune-UK, and for DeepFaune v1.5.0 the 0.5 threshold of its example script rather than the 0.8 interface default, which is the more favourable setting for DeepFaune on this set (Section 4.1).

\subsection{Out-of-Distribution Dataset}

The out-of-distribution (OOD) evaluation set is drawn from the MammalWeb citizen-science archive~\parencite{hsing2018,hsing2018data}, which was collected separately from the master dataset, although some of its images later proved to be in the master archive (see below). MammalWeb provides sequence-level (not per-image) expert species labels with no bounding boxes: 10,483 unique gold-standard sequences, of which 10,387 have photos listed in the sequence-to-photo mapping table (35,109 rows; 34,996 unique photos). After restricting to the 21 animal labels (excluding ``Nothing'' and ``Human'') and de-duplicating to one row per photo, 25,051 candidate photos from 7,100 sequences remained. Localisation was supplied by running MegaDetector v5a~\parencite{megadetector5} (confidence $\geq$ 0.2) over every candidate photo; each resulting animal box was assigned the photo's sequence-level species label, and photos with no qualifying detection were kept as empty images. The set was then restricted to the 14 species that also appear in the DeepFaune-UK vocabulary, giving 21,640 images.

Because these labels and boxes are only as reliable as the automated steps that produced them, every image was then reviewed by hand. Species labels were corrected (including images containing more than one species); of the 20,988 automatically generated boxes, 774 were removed as erroneous, 180 were added for missed animals and 225 were adjusted (196 by more than 0.5\% of the image dimensions). The reviewed set contains 14 species, 2,896 empty images and 47 images with more than one species. Because the boxes and labels were human-verified, the set is less dependent on any single detector, although it was seeded with MegaDetector v5a boxes (the detector used inside SpeciesNet), 95\% of which were kept unchanged. Apart from the detector comparisons described in Section 3.3, the OOD set was not deliberately used in developing DeepFaune-UK: not for training, classifier selection or threshold tuning. Some of its images had nevertheless entered the master dataset, as described next.

Because the master images are stored under the SHA-256 hash of their content (Section 3.1), the OOD images could be checked for exact copies in the master dataset. This check found 1,039 OOD images (4.8\%) byte-identical to master images, 954 of them in the training split, 40 in validation and 45 in test, most of them containing pheasant (611) or badger (210). They came from 440 of the 6,019 sequences in the 21,640-image set and from 25 of the 51 contributors, so imagery from some MammalWeb sites had entered the master archive. Every sequence containing such an image was therefore removed, leaving 20,212 images in 5,579 sequences (2,857 empty and 44 with more than one species), which form the evaluation set for all results reported here unless stated otherwise; scores on the original 21,640 images are given in Table B3. Because the remaining images of the overlapping contributors may come from the same cameras as training images, we also report results on a site-disjoint subset restricted to the 26 contributors none of whose images overlapped with the master dataset (1,850 images in 622 sequences, 243 of them empty, covering 12 of the 14 species but no muntjac or red squirrel); sequences without a recorded contributor are excluded from this subset because their site is unknown. Only byte-identical copies were detected; re-encoded or resized copies of the same photographs would not have been.

\subsection{Evaluation Protocol}

Because both baselines output a single label per image, all models were compared at the image level. For DeepFaune-UK, the image label is the species of the most confident animal detection (detector confidence $\geq$ 0.25), the detector's label if only people or vehicles are detected, and ``empty'' otherwise. Ground truth for each image is the set of species present, or ``empty''.

The models use different label vocabularies. DeepFaune names some taxa only at coarse levels (e.g., ``squirrel'', ``lagomorph'', ``equid'') and birds only in coarse groups such as ``columbiform'' and ``corvid''; SpeciesNet may return a genus, family or higher rank. We therefore report two levels:

\begin{itemize}
\item \textbf{Strict species level:} a prediction is correct only if it names the exact species.

\item \textbf{Like-for-like level:} a prediction is correct if it identifies the correct animal at the resolution that DeepFaune's vocabulary allows: any squirrel label is correct for either squirrel species, ``lagomorph'' for rabbit, ``equid'' for horse, ``corvid'' for carrion crow, ``galliform'' for pheasant and ``columbiform'' for woodpigeon. SpeciesNet predictions are mapped to these groups by the same principle: a label is credited when its taxon lies wholly within the group, namely genus \emph{Canis} (dog), \emph{Vulpes} (fox), \emph{Felis} (cat), \emph{Capreolus} (roe deer), \emph{Muntiacus} (muntjac), \emph{Equus} (horse), \emph{Meles} (badger) and \emph{Sciurus} (squirrel); family Erinaceidae (hedgehog), Corvidae (crow), Columbidae or order Columbiformes (woodpigeon) and Phasianidae or order Galliformes (pheasant); and order Lagomorpha (rabbit). Families that span more than one DeepFaune class (Sciuridae, which includes marmots; Canidae; Cervidae; Felidae; Mustelidae) are not credited. SpeciesNet species-level labels outside the 14 OOD species are mapped in the same way, so that, for example, European hare is credited for rabbit, wild turkey for pheasant and domestic donkey for horse (20 images in the evaluation set). DeepFaune-UK's labels are treated more conservatively: those outside the 14 OOD species are not mapped to DeepFaune's groups and always count as wrong, which affects a single image (a pheasant labelled capercaillie by v1.2). Muntjac (\emph{Muntiacus reevesi}), absent from DeepFaune's vocabulary, is retained and scored as zero for that model.

\end{itemize}
Formally, let $T_i$ be the set of true labels of image $i$ (one or more species, or empty), $\hat{y}_i$ the model's image label, and $g(\cdot)$ the mapping from a label to its like-for-like group, so that, for example, $g(\text{red squirrel}) = g(\text{grey squirrel}) = \text{squirrel}$. Image $i$ is scored as correct when
\begin{equation}
\hat{y}_i \in T_i \quad \text{(strict)} \qquad \text{or} \qquad g(\hat{y}_i) \in g(T_i) \quad \text{(like-for-like)},
\label{eq:correct}
\end{equation}
with $g(\text{empty}) = \text{empty}$. For per-class counts, each prediction is mapped to a species label $\tilde{y}_i$. At the strict level $\tilde{y}_i = \hat{y}_i$. At the like-for-like level a correct group label is credited to the species it matches, and a wrong group label is charged to a fixed representative species of that group, its more common member in the OOD set:
\begin{equation}
\tilde{y}_i = \begin{cases} c & \text{if } c \in T_i \text{ and } g(c) = g(\hat{y}_i), \\ r\big(g(\hat{y}_i)\big) & \text{otherwise,} \end{cases}
\label{eq:remap}
\end{equation}
where $r(\cdot)$ returns grey squirrel for ``squirrel'', the single member for the other groups that contain one of the 14 species, empty for empty, and a label outside the 14 species (which can only be a false positive) for any other group. For each class $c$, $\mathrm{TP}_c = |\{i : c \in T_i,\ \tilde{y}_i = c\}|$, $\mathrm{FN}_c = |\{i : c \in T_i,\ \tilde{y}_i \neq c\}|$ and $\mathrm{FP}_c = |\{i : c \notin T_i,\ \tilde{y}_i = c\}|$, giving precision $P_c = \mathrm{TP}_c / (\mathrm{TP}_c + \mathrm{FP}_c)$, recall $R_c = \mathrm{TP}_c / (\mathrm{TP}_c + \mathrm{FN}_c)$ and $\mathrm{F1}_c = 2 P_c R_c / (P_c + R_c)$, with any ratio whose denominator is zero taken as 0 (as for muntjac, which DeepFaune v1.5.0 never predicts). We report these for each species and summarise them with accuracy over all images and with
\begin{equation}
\text{macro-F1} = \frac{1}{|S|} \sum_{c \in S} \mathrm{F1}_c, \qquad \text{weighted F1} = \sum_{c \in S} \frac{n_c}{\sum_{k \in S} n_k} \, \mathrm{F1}_c,
\label{eq:f1}
\end{equation}
where $S$ is the set of species present in the evaluation set (all 14 in the main set) and $n_c = \mathrm{TP}_c + \mathrm{FN}_c$ is the number of images containing species $c$; macro-averaged precision and recall are defined in the same way. Empty images count towards accuracy and are reported as a separate class, but do not enter these averages. Images from the same camera-trap sequence are strongly correlated: the 20,212 evaluation images come from 5,579 MammalWeb sequences, and all three red-squirrel images, for example, belong to one sequence. Uncertainty was therefore estimated with a cluster bootstrap~\parencite{efron1993,field2007} that resamples whole sequences rather than single images (2,000 resamples), giving percentile 95\% confidence intervals. Paired differences in accuracy between two models were assessed with the same resamples: if $d^{*}_b$ is the difference in resample $b$ ($b = 1, \ldots, B$, $B = 2000$), the two-sided p-value is $p = \min\{1,\ 2 \min(\#\{d^{*}_b \leq 0\}, \#\{d^{*}_b \geq 0\}) / B\}$, reported as $p < 0.001$ when no resample reaches zero. In a resample that contains no images of a species, macro-F1 is averaged over the species that are present rather than scoring the missing species as zero; this happens in 35--37\% of resamples for each of muntjac and red squirrel, whose three images each come from a single sequence. As a stricter check on generalisation across sites, the analysis was repeated with clusters defined by the anonymised MammalWeb contributor who uploaded each sequence (814 clusters: 51 contributors, plus the sequences without a recorded contributor, 11\% of images, each treated as its own cluster), which approximates resampling cameras or sites. For comparison with common practice we also report McNemar's test~\parencite{mcnemar1947} on per-image correctness, using the continuity-corrected statistic~\parencite{edwards1948}
\begin{equation}
\chi^2 = \frac{\left( |n_{10} - n_{01}| - 1 \right)^2}{n_{10} + n_{01}},
\label{eq:mcnemar}
\end{equation}
where $n_{10}$ is the number of images that the first model labels correctly and the second does not, $n_{01}$ is the number for which the reverse holds, and $\chi^2$ is referred to a $\chi^2$ distribution with one degree of freedom. Because McNemar's test treats images as independent, it overstates the evidence when images are clustered, and we report it only as a secondary statistic.

To check whether differences arise from operating thresholds rather than from the models themselves, we also selected, for each DeepFaune-UK version, a detector threshold maximising class-agnostic box-level F1 (IoU $\geq$ 0.5) and a classifier confidence threshold, below which the image is labelled ``undefined'' and counted as wrong, maximising image-level macro-F1 over the 28 classes, both on the master validation split (4,626 images, none empty), and applied them unchanged to the OOD set.

\subsection{Learning Curve and Resource Footprint}

To address RQ2, the DeepFaune-UK classifier was retrained on stratified random subsets of 10\%, 25\%, 50\% and 100\% of the training crops, each with three random seeds (12 runs), using the identical recipe of Section 3.3; the validation and test splits were left unchanged. Subsets were drawn independently within each class, so that class proportions are preserved up to rounding (at least one crop per class). Each model was evaluated on the in-distribution test crops and on the OOD set, using the same MegaDetector outputs for every model so that only the classifier varies. For each run we recorded wall-clock training time, peak GPU memory allocated by PyTorch, and GPU board energy, obtained by sampling power draw with nvidia-smi at 1 Hz during training and summing the 1-second samples; time, memory and energy exclude the evaluation pass. Carbon emissions were estimated from energy using the 2024 UK conversion factor for electricity generated (0.207 kg CO\textsubscript{2}e kWh$^{-1}$, excluding transmission and distribution losses)~\parencite{desnz2024}. To separate the effect of the starting checkpoint from that of training-set size, the 10\% and 100\% conditions were repeated with the DINOv3-based classifier of Section 3.3 (three seeds each, six runs), with OOD crops resized to that model's 224-pixel input.

\section{Results}

\subsection{Out-of-Distribution Comparison}

At the exact-species level, DeepFaune-UK v1.2 labelled 88.8\% of the 20,212 OOD images correctly, counting the 2,857 empty images, which are correct when no animal is reported (95\% CI 88.0--89.5\%; macro-F1 0.778), compared with 59.0\% for SpeciesNet (macro-F1 0.536) and 56.5\% for DeepFaune v1.5.0 (macro-F1 0.398) (Table 1). This gap is expected and is largely a matter of vocabulary rather than recognition: DeepFaune has no labels for grey or red squirrel, European rabbit, carrion crow, woodpigeon, pheasant, horse or muntjac (only ``squirrel'', ``lagomorph'', ``equid'' and coarse bird groups; see examples in Figure A1), and SpeciesNet frequently returned family-level or coarser labels --- for example ``rabbit and hare family'' (2,609 images), ``cervidae family'' (2,264), ``animal'' (734), ``bird'' (653) and ``mammal'' (506). The strict comparison therefore measures whether a model delivers the labels a UK user needs, not whether it recognises the animal.

When predictions were scored at the resolution DeepFaune's vocabulary allows (Section 3.6), DeepFaune v1.5.0 achieved the highest accuracy (91.4\%, 95\% CI 90.9--92.0\%), followed by DeepFaune-UK v1.2 (88.8\%, 88.1--89.6\%) and SpeciesNet (73.4\%, 72.0--74.8\%) (Table 1). Macro-F1 was almost identical for DeepFaune v1.5.0 and DeepFaune-UK v1.2 (0.855 vs. 0.850); DeepFaune-UK had higher macro recall (0.866 vs. 0.829) and DeepFaune higher macro precision (0.886 vs. 0.844). SpeciesNet had the highest precision (0.911) but the lowest recall (0.605), reflecting its tendency to back off to higher taxonomic ranks when uncertain. In paired comparisons, DeepFaune v1.5.0 was correct on 1,436 images where DeepFaune-UK was wrong, against 918 in the opposite direction (difference 2.6 percentage points, 95\% CI 1.8--3.3, p $<$ 0.001; McNemar $\chi^{2}$ = 113.5), and both of these models outperformed SpeciesNet (p $<$ 0.001 under both clusterings). With images clustered by contributor, the interval for the difference between DeepFaune v1.5.0 and DeepFaune-UK v1.2 widened to $-$0.2 to 4.6 points (p = 0.08), so the advantage of DeepFaune v1.5.0 is clear across sequences but less certain across contributors' sites. Macro-F1 confidence intervals are wide and asymmetric (Tables B1 and B2) because muntjac and red squirrel have only three images each, each set from a single sequence, so each species is absent from many resamples and, when present, contributes an extreme F1 (0 for muntjac at both levels; 0 for red squirrel at the strict level and 1.0 for most models at the like-for-like level). At the 0.8 threshold of DeepFaune's graphical interface, DeepFaune v1.5.0 would have scored 89.5\% like-for-like (55.5\% strict), because more images would be labelled ``undefined''; the comparisons above use its more favourable 0.5 threshold.

\begin{table}[htbp]
\centering
\caption{Image-level performance on the OOD evaluation set (20,212 images, 14 species plus empty). DeepFaune-UK v1.3 is described in Section 4.3. Best value per column in bold; sequence- and contributor-clustered 95\% bootstrap confidence intervals for accuracy and macro-F1 are given in Tables B1 and B2.}
\label{tab:1}
\footnotesize\setlength{\tabcolsep}{4pt}
\begin{tabularx}{\textwidth}{>{\raggedright\arraybackslash}p{2.3cm}>{\centering\arraybackslash}X>{\centering\arraybackslash}X>{\centering\arraybackslash}X>{\centering\arraybackslash}X>{\centering\arraybackslash}X>{\centering\arraybackslash}X>{\centering\arraybackslash}X}
\toprule
\textbf{Model} & \textbf{Strict accuracy} & \textbf{Strict macro-F1} & \textbf{Like-for-like accuracy} & \textbf{Like-for-like macro-P} & \textbf{Like-for-like macro-R} & \textbf{Like-for-like macro-F1} & \textbf{Like-for-like weighted F1} \\
\midrule
DeepFaune-UK v1.2 & 0.888 & 0.778 & 0.888 & 0.844 & 0.866 & 0.850 & 0.928 \\
DeepFaune-UK v1.3 & \textbf{0.906} & \textbf{0.786} & 0.906 & 0.844 & \textbf{0.880} & \textbf{0.857} & 0.942 \\
DeepFaune v1.5.0 & 0.565 & 0.398 & \textbf{0.914} & 0.886 & 0.829 & 0.855 & \textbf{0.947} \\
SpeciesNet 5.0.5 & 0.590 & 0.536 & 0.734 & \textbf{0.911} & 0.605 & 0.708 & 0.809 \\
\bottomrule
\end{tabularx}
\end{table}
Per-species F1 scores (Table 2, Figure 2) show that the models' strengths differ. DeepFaune-UK v1.2 outperformed DeepFaune v1.5.0 for grey squirrel (0.962 vs. 0.923), hedgehog (0.901 vs. 0.827), woodpigeon (0.978 vs. 0.963) and carrion crow (0.831 vs. 0.684). DeepFaune v1.5.0 had higher F1 for the other eight species with more than three images, most clearly roe deer (0.968 vs. 0.926), cat (0.954 vs. 0.763), dog (0.903 vs. 0.847) and red fox (0.945 vs. 0.924), and was more reliable on empty images (F1 0.847 vs. 0.803). SpeciesNet was best on empty images (F1 0.920) and close to the others for rabbit, grey squirrel and dog, and better than DeepFaune-UK for cat, but its recall was low for woodpigeon (0.22), carrion crow (0.38), horse (0.48), roe deer (0.51), hedgehog (0.53) and pheasant (0.60). Figures A2 and A3 show examples of difficult images, often taken at night or showing only part of the animal, on which neither baseline returned the correct group while both DeepFaune-UK models named the exact species; in two of these DeepFaune v1.5.0 named a species absent from Britain (raccoon and chamois). No model identified muntjac: it is not in DeepFaune's vocabulary, and DeepFaune-UK v1.2 assigned 91 images to muntjac (v1.3: 242) but neither version gave that label to any of the three true muntjac images.

\begin{table}[htbp]
\centering
\caption{Like-for-like per-species F1 on the OOD evaluation set. n = number of images containing the species; best value per row in bold (ties at three decimals are all bolded; no bold where all models score 0). At this level any squirrel label counts as correct for either squirrel species, so the red-squirrel values (three images, all from one sequence) do not show that red squirrels were identified: at the strict species level, no model labelled any of them as red squirrel (F1 = 0 for all four).}
\label{tab:2}
\footnotesize\setlength{\tabcolsep}{4pt}
\begin{tabularx}{\textwidth}{>{\raggedright\arraybackslash}p{2.3cm}>{\centering\arraybackslash}X>{\centering\arraybackslash}X>{\centering\arraybackslash}X>{\centering\arraybackslash}X>{\centering\arraybackslash}X}
\toprule
\textbf{Species} & \textbf{n} & \textbf{DeepFaune-UK v1.2} & \textbf{DeepFaune-UK v1.3} & \textbf{DeepFaune v1.5.0} & \textbf{SpeciesNet 5.0.5} \\
\midrule
Roe deer & 5,361 & 0.926 & 0.940 & \textbf{0.968} & 0.674 \\
Rabbit & 3,584 & 0.939 & \textbf{0.958} & 0.950 & 0.934 \\
Grey squirrel & 2,742 & \textbf{0.962} & 0.960 & 0.923 & 0.960 \\
Badger & 1,448 & 0.959 & \textbf{0.973} & 0.971 & 0.907 \\
Red fox & 920 & 0.924 & \textbf{0.946} & 0.945 & 0.897 \\
Cat & 843 & 0.763 & 0.840 & \textbf{0.954} & 0.928 \\
Woodpigeon & 673 & 0.978 & \textbf{0.979} & 0.963 & 0.361 \\
Hedgehog & 670 & \textbf{0.901} & 0.889 & 0.827 & 0.685 \\
Pheasant & 513 & 0.916 & 0.875 & \textbf{0.922} & 0.742 \\
Dog & 324 & 0.847 & 0.870 & \textbf{0.903} & 0.847 \\
Horse & 278 & 0.948 & \textbf{0.966} & 0.961 & 0.646 \\
Carrion crow & 37 & \textbf{0.831} & 0.804 & 0.684 & 0.528 \\
Muntjac & 3 & 0.000 & 0.000 & 0.000 & 0.000 \\
Red squirrel & 3 & \textbf{1.000} & \textbf{1.000} & \textbf{1.000} & 0.800 \\
Empty & 2,857 & 0.803 & 0.803 & 0.847 & \textbf{0.920} \\
\bottomrule
\end{tabularx}
\end{table}
\begin{figure}[htbp]
\centering
\includegraphics[width=\textwidth]{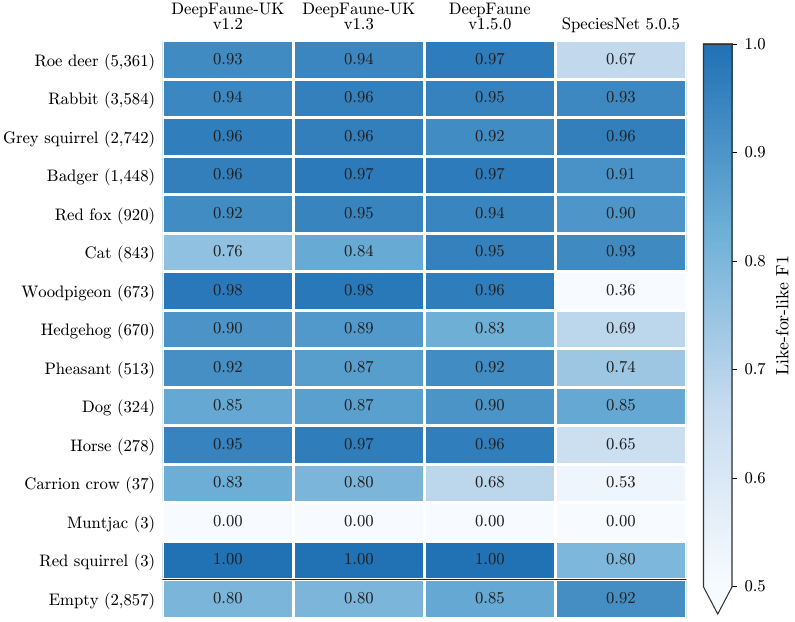}
\caption{Like-for-like per-species F1 on the OOD evaluation set for the four models. Species are ordered by the number of images containing them (in brackets); cells are shaded by F1, with values below 0.5 sharing the lightest shade. Muntjac and red squirrel have only three images each, each set from a single sequence, so their values are unreliable; the red-squirrel values reflect the shared squirrel group, and no model identified a red squirrel at the strict species level.}
\label{fig:2}
\end{figure}

\begin{figure}[htbp]
\centering
\includegraphics[width=\textwidth]{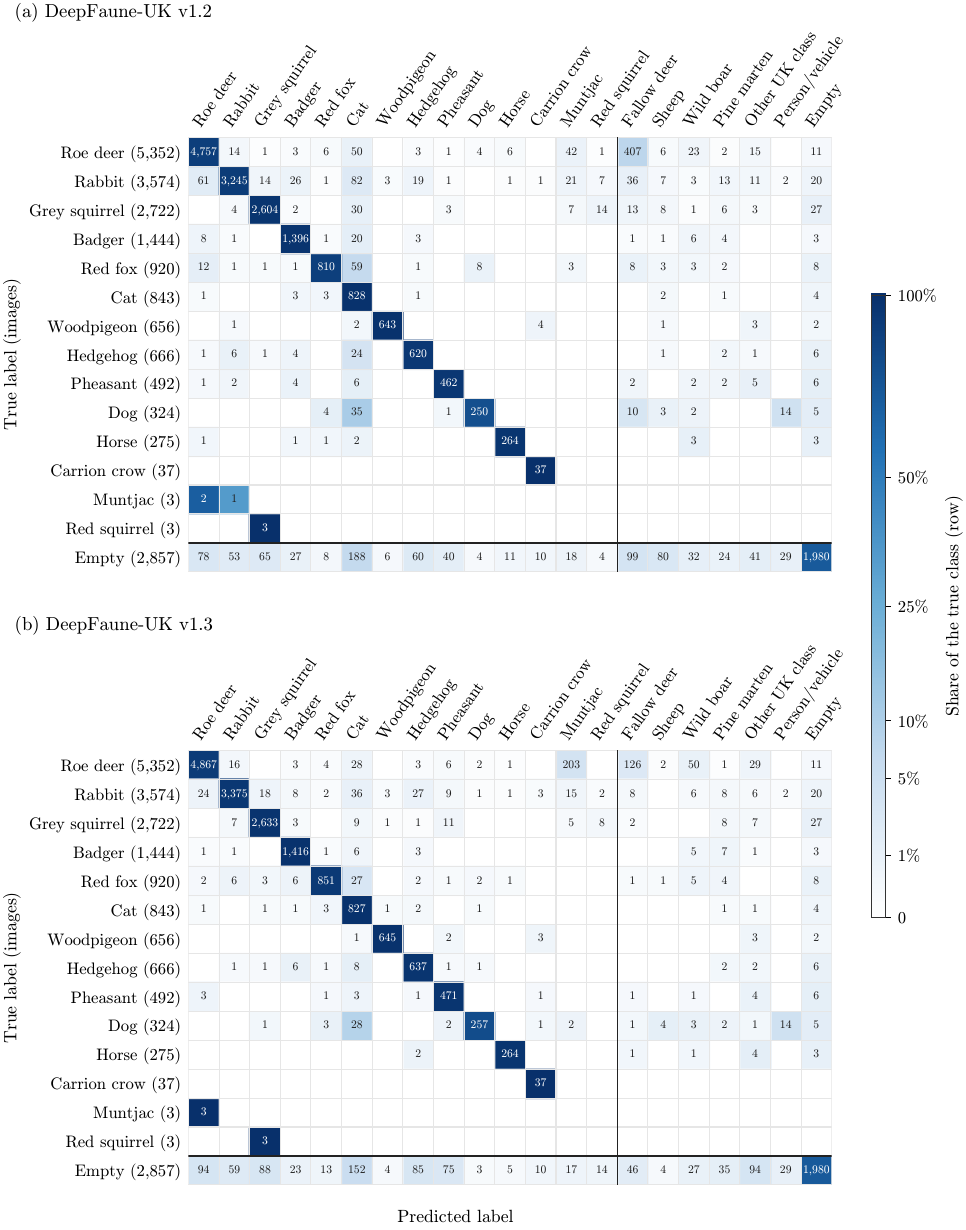}
\caption{Confusion matrices of (a) DeepFaune-UK v1.2 and (b) DeepFaune-UK v1.3 on the OOD evaluation set at the strict species level (20,168 single-species and empty images; the 44 images with more than one species are excluded). Numbers are image counts, and shading shows each count as a share of its true class on a square-root scale. Columns to the right of the vertical rule are predicted labels outside the 14 OOD species: fallow deer, sheep, wild boar, pine marten, the remaining UK classes (pooled), person or vehicle, and empty.}
\label{fig:3}
\end{figure}

Two error types account for most of the gap between DeepFaune-UK v1.2 and DeepFaune v1.5.0 (Figure 3a). First, DeepFaune-UK labelled 407 roe deer images (7.6\%) as fallow deer (\emph{Dama dama}), a species absent from the OOD set but present in its training vocabulary (Figure A4a); DeepFaune, in contrast, rarely invoked fallow deer (6 images). Second, DeepFaune-UK assigned an animal label to 848 of the 2,857 empty images (29.7\%, against 11.3\% for DeepFaune and 7.4\% for SpeciesNet), most often cat (188), fallow deer (99) and sheep (80) (Figure A5d); together with misidentifications of other species, this gave cat a precision of only 0.62. Both are errors of a model with a larger UK vocabulary meeting images from other sites: the extra classes give it more ways to be wrong, and detector false positives are passed to the classifier rather than rejected. Most of these empty-frame detections are on background, but in a visual check of a random sample of 48 of them at least five showed an animal that the reference labels had missed (grey squirrels, a rabbit and blackbirds), so a small part of this error lies in the reference labels rather than in the model. For animal presence, DeepFaune-UK had high recall (99.4\% of animal images given an animal label, against 96.5\% for DeepFaune and 98.2\% for SpeciesNet, counting DeepFaune's ``undefined'' as an animal label and SpeciesNet's ``no cv result'' as not). Most of the remaining 0.6\% (95 of 111 images; the rest were labelled person) received no detection above the 0.25 threshold, so they never reached the classifier, even when clearly visible (Figure A5c).

\subsection{Operating Thresholds}

To test whether these differences were an artefact of default thresholds, DeepFaune-UK's thresholds were selected on the master validation split (4,626 images). Class-agnostic box-level F1 (IoU $\geq$ 0.5) peaked at a detector confidence of 0.50 (F1 0.943, against 0.929 at the default of 0.25) and was essentially flat between 0.40 and 0.55. Image-level macro-F1 was insensitive to the classifier confidence threshold up to about 0.8 (0.958, to three decimals, at every tested threshold from 0.0 to 0.50); the maximum was shared by 0.30 and 0.35, and the lower, 0.30, was selected. Applied unchanged to the OOD set, the calibrated operating point gave like-for-like accuracy of 0.889 and macro-F1 of 0.837 (default: 0.888 and 0.850). Fewer empty images were labelled as animals (empty F1 0.824 vs. 0.803) and cat improved (F1 0.837 vs. 0.763), but these gains were offset by lost recall elsewhere (macro recall 0.813 vs. 0.866), and the difference from DeepFaune v1.5.0 remained (2.6 percentage points; p $<$ 0.001 with sequence clustering and p = 0.07 with contributor clustering; McNemar $\chi^{2}$ = 127.8). The gap between the two DeepFaune-based models therefore lies in the models' errors on the OOD images (classifier confusions and detector false positives on empty frames) rather than in the choice of thresholds.

\subsection{Following the Upstream Model: DeepFaune-UK v1.3}

Applying the identical fine-tuning recipe to the DINOv3-based DeepFaune v1.5.0 classifier produced DeepFaune-UK v1.3, which improved on v1.2 at both label levels (Tables 1 and 2). Strict accuracy rose from 88.8\% to 90.6\% (95\% CI 89.9--91.3\%; macro-F1 0.786), and like-for-like accuracy from 88.8\% to 90.6\% (89.9--91.3\%), with like-for-like macro-F1 of 0.857 against 0.850. In paired like-for-like comparisons, v1.3 was correct on 659 images where v1.2 was wrong, against 301 in the opposite direction (difference 1.8 percentage points, 95\% CI 1.4--2.2, p $<$ 0.001, and p $<$ 0.001 with contributor clustering; McNemar $\chi^{2}$ = 132.8). Against DeepFaune v1.5.0, the difference in like-for-like accuracy narrowed from 2.6 to 0.8 percentage points (936 against 1,096 discordant images). This residual difference was significant when images were resampled by sequence (95\% CI 0.1--1.5 points, p = 0.02; McNemar $\chi^{2}$ = 12.4) but not when they were resampled by contributor (95\% CI $-$1.6 to 2.7 points, p = 0.53), so there is no clear evidence that it would persist across new sites. DeepFaune-UK v1.3 had the highest like-for-like macro-F1 and macro recall of the four models (0.857 and 0.880), although its sequence-clustered macro-F1 confidence interval (0.837--0.929) overlaps that of DeepFaune v1.5.0 (0.829--0.930; Table B2).

The gain came mainly from the classifier errors identified in Section 4.1 (Figure 3b). Roe deer labelled as fallow deer fell from 407 to 126 images (Figure A4a), raising roe deer F1 from 0.926 to 0.940, and cat F1 rose from 0.763 to 0.840 as fewer animals of other species were labelled cat (310 against 146; Figure A4c,d), although both remained below DeepFaune v1.5.0 (0.968 and 0.954). In contrast, exactly the same 848 empty images received an animal label as with v1.2, although the labels changed (v1.3 called fewer of them sheep or fallow deer), so empty-image F1 was unchanged (0.803): these errors begin with the detector, and a classifier without an empty class cannot reject them (Figure A5d). v1.3 also assigned more images to muntjac (242 against 91), again with none correct; 203 of these were roe deer (Figure A5a). The squirrel results were similar to those of v1.2: all three red squirrel images were labelled grey (Figure A5b), while the number of grey squirrel images labelled red fell from 14 to 8, although v1.3 labelled 14 empty images red squirrel (4 for v1.2). Recalibrating v1.3's thresholds on the validation split as in Section 4.2 selected a detector threshold of 0.50 and a classifier threshold of 0.20, and gave like-for-like accuracy of 0.909 and macro-F1 of 0.858 on the OOD set; the remaining difference from DeepFaune v1.5.0 (0.5 percentage points) was not significant with sequence clustering (p = 0.08; p = 0.70 with contributor clustering), although the unclustered McNemar test, which ignores within-sequence correlation, was ($\chi^{2}$ = 6.8, p = 0.009).

\subsection{Learning Curve}

Performance on the OOD set was essentially independent of training-set size over the range tested (Figure 4a--c; Table 3). With 10\% of the training crops (4,270 crops, about 150 per class on average), like-for-like macro-F1 was 0.838 $\pm$ 0.007 (mean $\pm$ s.d., three seeds), against 0.839 $\pm$ 0.009 with all 42,719 crops; the 25\% models were nominally the best (0.843 $\pm$ 0.006). Across the four fractions the means differed by at most 0.006 in like-for-like macro-F1, 0.008 in like-for-like accuracy and 0.006 in strict-species macro-F1 (range 0.767--0.773), comparable to seed-to-seed variation (s.d. up to 0.018); with three seeds, this shows the absence of a detectable trend rather than formal equivalence. In-distribution test-crop macro-F1 was high at every fraction (0.990 at 10\%, 0.993 at 100\%). The full-data retraining came close to the released DeepFaune-UK v1.2 model, although the released model scored slightly above all three retrained seeds (like-for-like accuracy 0.879 $\pm$ 0.005, range 0.875--0.885, vs. 0.888).

\begin{table}[htbp]
\centering
\caption{Learning curve and resource use (mean $\pm$ s.d. over three seeds; single NVIDIA RTX A6000). OOD scores are image-level on the evaluation set; detector outputs are identical for all runs.}
\label{tab:3}
\footnotesize\setlength{\tabcolsep}{4pt}
\begin{tabularx}{\textwidth}{>{\raggedright\arraybackslash}p{2.3cm}>{\centering\arraybackslash}X>{\centering\arraybackslash}X>{\centering\arraybackslash}X>{\centering\arraybackslash}X>{\centering\arraybackslash}X>{\centering\arraybackslash}X>{\centering\arraybackslash}X>{\centering\arraybackslash}X}
\toprule
\textbf{Training fraction} & \textbf{Training crops} & \textbf{ID test macro-F1} & \textbf{OOD strict macro-F1} & \textbf{OOD like-for-like accuracy} & \textbf{OOD like-for-like macro-F1} & \textbf{Training time (min)} & \textbf{Peak GPU memory (GB)} & \textbf{GPU energy (kWh)} \\
\midrule
10\% & 4,270 & 0.990 $\pm$ 0.002 & 0.767 $\pm$ 0.008 & 0.887 $\pm$ 0.005 & 0.838 $\pm$ 0.007 & 21.8 $\pm$ 0.9 & 3.8 & 0.064 $\pm$ 0.006 \\
25\% & 10,681 & 0.992 $\pm$ $<$0.001 & 0.773 $\pm$ 0.005 & 0.888 $\pm$ 0.010 & 0.843 $\pm$ 0.006 & 20.6 $\pm$ 3.9 & 3.8 & 0.071 $\pm$ 0.011 \\
50\% & 21,360 & 0.990 $\pm$ 0.003 & 0.767 $\pm$ 0.009 & 0.882 $\pm$ 0.018 & 0.838 $\pm$ 0.009 & 30.8 $\pm$ 5.5 & 3.8 & 0.122 $\pm$ 0.021 \\
100\% & 42,719 & 0.993 $\pm$ 0.002 & 0.767 $\pm$ 0.009 & 0.879 $\pm$ 0.005 & 0.839 $\pm$ 0.009 & 44.3 $\pm$ 7.6 & 3.8 & 0.192 $\pm$ 0.034 \\
\bottomrule
\end{tabularx}
\end{table}
With the DINOv3 backbone (Table 4; open squares in Figure 4), accuracy showed the same pattern: like-for-like accuracy was 0.901 $\pm$ 0.013 with 10\% of the training crops and 0.903 $\pm$ 0.003 with all of them, but like-for-like macro-F1 was lower at 10\% (0.846 $\pm$ 0.005 vs. 0.857 $\pm$ 0.003). At both fractions, the DINOv3 models were more accurate than their DINOv2 counterparts, by 1.5 and 2.4 percentage points of like-for-like accuracy at 10\% and 100\% respectively, and by 0.008 and 0.018 in macro-F1, so on the main evaluation set the choice of starting checkpoint mattered more than a tenfold change in training data; on the site-disjoint subset it made no difference (Section 4.6). The released DeepFaune-UK v1.3 is the full-data seed-2 model, the best of the three on validation macro-F1 (0.99431, against 0.99420 and 0.99429); it was also the most accurate of the three on the OOD set (like-for-like accuracy 0.906, range 0.901--0.906), although the margins between seeds are small on both sets.

\begin{table}[htbp]
\centering
\caption{Backbone comparison: DINOv2 (DeepFaune v1.4, as in Table 3) and DINOv3 (DeepFaune v1.5.0) classifiers fine-tuned with the identical recipe (mean $\pm$ s.d. over three seeds).}
\label{tab:4}
\footnotesize\setlength{\tabcolsep}{4pt}
\begin{tabularx}{\textwidth}{>{\raggedright\arraybackslash}p{2.3cm}>{\centering\arraybackslash}X>{\centering\arraybackslash}X>{\centering\arraybackslash}X>{\centering\arraybackslash}X>{\centering\arraybackslash}X>{\centering\arraybackslash}X>{\centering\arraybackslash}X>{\centering\arraybackslash}X}
\toprule
\textbf{Backbone} & \textbf{Training fraction} & \textbf{ID test macro-F1} & \textbf{OOD strict macro-F1} & \textbf{OOD like-for-like accuracy} & \textbf{OOD like-for-like macro-F1} & \textbf{Training time (min)} & \textbf{Peak GPU memory (GB)} & \textbf{GPU energy (kWh)} \\
\midrule
DINOv2 & 10\% & 0.990 $\pm$ 0.002 & 0.767 $\pm$ 0.008 & 0.887 $\pm$ 0.005 & 0.838 $\pm$ 0.007 & 21.8 $\pm$ 0.9 & 3.8 & 0.064 $\pm$ 0.006 \\
DINOv2 & 100\% & 0.993 $\pm$ 0.002 & 0.767 $\pm$ 0.009 & 0.879 $\pm$ 0.005 & 0.839 $\pm$ 0.009 & 44.3 $\pm$ 7.6 & 3.8 & 0.192 $\pm$ 0.034 \\
DINOv3 & 10\% & 0.991 $\pm$ 0.002 & 0.779 $\pm$ 0.005 & 0.901 $\pm$ 0.013 & 0.846 $\pm$ 0.005 & 45.6 $\pm$ 14.7 & 4.4 & 0.114 $\pm$ 0.014 \\
DINOv3 & 100\% & 0.994 $\pm$ $<$0.001 & 0.786 $\pm$ 0.003 & 0.903 $\pm$ 0.003 & 0.857 $\pm$ 0.003 & 91.2 $\pm$ 20.6 & 4.4 & 0.394 $\pm$ 0.088 \\
\bottomrule
\end{tabularx}
\end{table}
\begin{figure}[htbp]
\centering
\includegraphics[width=\textwidth]{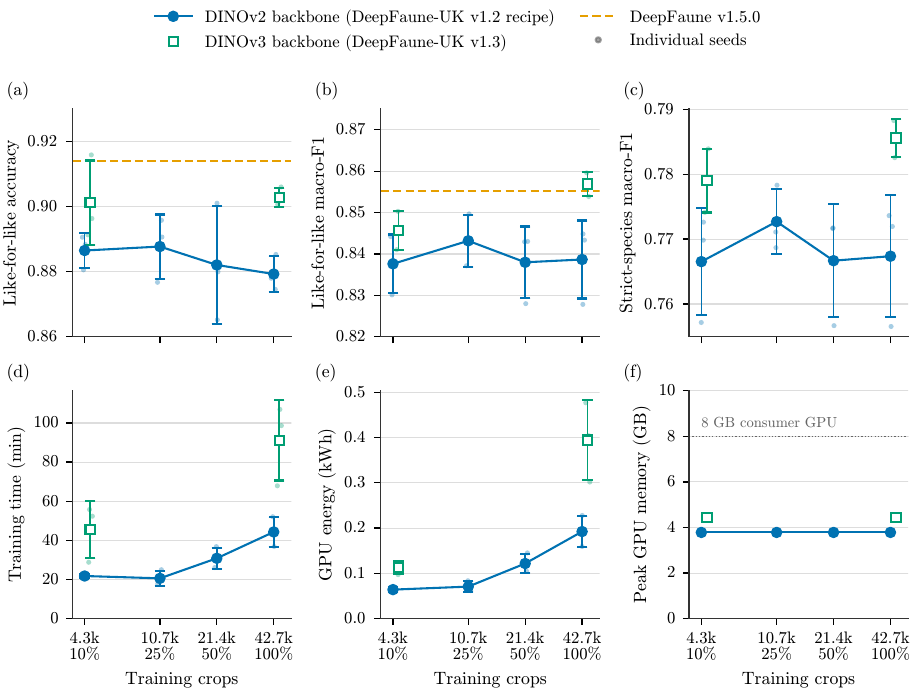}
\caption{Learning curve and resource footprint of regional fine-tuning (mean $\pm$ s.d. over three seeds; small points show individual seeds). (a--c) OOD performance against the number of training crops: (a) like-for-like accuracy, (b) like-for-like macro-F1 and (c) strict-species macro-F1; in (a) and (b) the dashed line marks DeepFaune v1.5.0, and SpeciesNet (like-for-like accuracy 0.734, macro-F1 0.708) lies below the plotted range; both baselines fall well below the range of (c) (strict macro-F1 0.398 and 0.536). (d--f) Training time, GPU energy and peak GPU memory per run on a single NVIDIA RTX A6000; the dotted line in (f) marks the 8 GB memory of a typical consumer GPU. Filled circles: DINOv2 backbone (DeepFaune-UK v1.2 recipe); open squares: DINOv3 backbone (DeepFaune-UK v1.3), trained at 10\% and 100\% only and offset slightly for legibility.}
\label{fig:4}
\end{figure}

\subsection{Resource Footprint}

Every DINOv2 run fitted in 3.8 GB of GPU memory (peak memory allocated by PyTorch on the 48 GB card, excluding the CUDA context and allocator overhead; Figure 4f), which suggests that training would fit on current 8--12 GB consumer graphics cards, although we did not test such hardware. A full-data training run took 44 $\pm$ 8 minutes and consumed 0.19 $\pm$ 0.03 kWh of GPU energy; a 10\% run took 22 minutes and 0.06 kWh. Training time did not scale linearly with data because smaller training sets ran for more epochs before early stopping (30--38 epochs at 10\% vs. 17--24 at 100\%), and each epoch includes a fixed-cost validation pass (5,410 crops) and the saving of a checkpoint. Mean GPU board power was 167--262 W, below the card's 300 W limit, indicating that the GPU was not fully utilised. The complete learning-curve experiment --- twelve training runs --- took 5.9 GPU-hours and 1.35 kWh, corresponding to approximately 0.28 kg CO\textsubscript{2}e at a UK grid factor of 0.207 kg CO\textsubscript{2}e kWh$^{-1}$~\parencite{desnz2024}; a single full-data run corresponds to about 0.04 kg CO\textsubscript{2}e. The DINOv3 backbone was more expensive to adapt. Every run peaked at 4.4 GB of allocated GPU memory, also below 8 GB; a full-data run took 91 $\pm$ 21 minutes and 0.39 $\pm$ 0.09 kWh, and a 10\% run 46 $\pm$ 15 minutes and 0.11 kWh, partly because the DINOv3 runs continued for more epochs (22--34 at 100\%, 37--45 at 10\%, one 10\% run reaching the 45-epoch limit without early stopping). The six DINOv3 runs together took 6.8 GPU-hours and 1.52 kWh (about 0.32 kg CO\textsubscript{2}e). These figures cover GPU energy only, exclude the host computer, and --- importantly --- exclude the cost of pre-training the backbone, the detector and the baselines, which was borne by their developers.

\subsection{Site-Disjoint Subset}

On the 1,850 images from the 26 contributors with no overlapping images (Section 3.5; Table 5), the like-for-like advantage of DeepFaune v1.5.0 was not seen: DeepFaune-UK v1.2 scored 90.5\%, DeepFaune-UK v1.3 90.6\% and DeepFaune v1.5.0 89.9\%, and none of the pairwise differences was significant (v1.2 against DeepFaune v1.5.0: +0.5 points, 95\% CI $-$1.2 to 2.3 with sequence clustering, p = 0.54; $-$1.9 to 4.5 with contributor clustering, p = 0.68; v1.3 against v1.2: +0.1 points, p = 0.83). SpeciesNet remained well below the DeepFaune-based models (79.1\%, p $<$ 0.001 against v1.2 with sequence clustering), and the strict-level ordering was unchanged. In the learning curve on this subset, DINOv3 models were no more accurate than DINOv2 models (like-for-like accuracy 0.901 vs. 0.900 at 10\% and 0.904 vs. 0.903 at 100\%), so the checkpoint effect seen on the main evaluation set did not appear here. The subset is small (9\% of the evaluation images, from 26 of the 51 contributors) and lacks the two rarest species, so its intervals are wide (Table 5); it gives no sign that DeepFaune-UK's results on the main set are inflated by sites shared with the training data, but it cannot resolve differences of one or two percentage points.

\begin{table}[htbp]
\centering
\caption{Site-disjoint subset: 1,850 images from the 26 contributors none of whose images were found as exact copies in the master dataset (622 sequences, 12 species plus empty; macro-F1 over the 12 species present). Confidence intervals from 2,000 cluster-bootstrap resamples.}
\label{tab:5}
\footnotesize\setlength{\tabcolsep}{4pt}
\begin{tabularx}{\textwidth}{>{\raggedright\arraybackslash}p{2.3cm}>{\centering\arraybackslash}X>{\centering\arraybackslash}X>{\centering\arraybackslash}X>{\centering\arraybackslash}X>{\centering\arraybackslash}X}
\toprule
\textbf{Model} & \textbf{Strict accuracy} & \textbf{Like-for-like accuracy} & \textbf{95\% CI, sequences} & \textbf{95\% CI, contributors} & \textbf{Like-for-like macro-F1} \\
\midrule
DeepFaune-UK v1.2 & 0.903 & 0.905 & 0.887--0.922 & 0.868--0.932 & 0.931 \\
DeepFaune-UK v1.3 & 0.906 & 0.906 & 0.888--0.923 & 0.863--0.934 & 0.904 \\
DeepFaune v1.5.0 & 0.515 & 0.899 & 0.880--0.917 & 0.843--0.927 & 0.877 \\
SpeciesNet 5.0.5 & 0.616 & 0.791 & 0.764--0.818 & 0.714--0.858 & 0.687 \\
\bottomrule
\end{tabularx}
\end{table}
\section{Discussion}

On out-of-distribution UK images, the regionally fine-tuned model did not outperform the larger model from which it was derived. At the label resolution DeepFaune itself supports, DeepFaune v1.5.0 was 2.6 percentage points more accurate than DeepFaune-UK v1.2 on the main evaluation set, with near-identical macro-F1, and recalibrating DeepFaune-UK's thresholds did not remove the difference. This is the result a sceptical reader would look for first, and we regard it as the honest headline of the benchmark. It should, however, be read in context. The advantage was uncertain across contributors and was not seen on the smaller site-disjoint subset (Section 4.6). DeepFaune v1.5.0 also benefits from a newer backbone, DINOv3 rather than DINOv2, and possibly from additional training data (it adds two new classes), so the comparison is between a 2026 continental model and a regional adaptation of its v1.4 predecessor rather than between two models built from the same starting point. DeepFaune-UK v1.3 tests this directly. When the identical recipe was applied to the v1.5.0 classifier, on the main evaluation set like-for-like accuracy rose by 1.8 percentage points, the gap to DeepFaune v1.5.0 shrank to 0.8 points, and macro-F1 became nominally the highest of any model, while the three-seed comparison on that set showed the effect of the starting checkpoint to be larger than the effect of a tenfold change in training data. On the main evaluation set, changing the starting checkpoint under the same regional training recipe therefore removed most of the observed accuracy gap; on the site-disjoint subset there was no significant gap to remove and no checkpoint effect. The v1.5.0 checkpoint differs from v1.4 in its backbone (DINOv3 rather than DINOv2, with 16- rather than 14-pixel patches), its input resolution (224 rather than 182 pixels) and any additional data on which DeepFaune trained it; these factors cannot be separated here, so the improvement should not be attributed to the DINOv3 backbone alone. A small residual gap of 0.8 percentage points remains on the main set, concentrated in empty frames (550 more images correct for DeepFaune v1.5.0, net) and roe deer (205), and, in F1 terms, in cats, a label DeepFaune-UK over-predicts. It is significant when images are resampled by sequence but not when they are resampled by contributor (Section 4.3), so it may not hold across new sites; to the extent that it is real, it plausibly reflects the larger and more diverse data behind DeepFaune's own classifier and its different detector configuration.

What regional fine-tuning did deliver was vocabulary. DeepFaune-UK achieved a strict image-level accuracy of 88.8\% (90.6\% for v1.3), including correct identification of empty frames, compared with 56.5\% and 59.0\% for the two baselines, while remaining within a few points of DeepFaune on the coarse groups the models share. It separates grey from red squirrel, with a test-crop F1 of 0.99 for both species at the training sites, and names UK birds at species level, distinctions that matter for UK monitoring and management. DeepFaune v1.5.0 cannot make them, because its vocabulary has a single squirrel class and assigns birds only to coarse groups; SpeciesNet has labels for both squirrel species and for these birds, but on this benchmark it often returned a genus, family or broader label instead (Section 4.1). DeepFaune-UK had a higher like-for-like F1 than both baselines for several species typical of UK deployments, clearly so for hedgehog, carrion crow and woodpigeon, and narrowly for grey squirrel, where SpeciesNet was almost as good. Its principal weaknesses were the confusion of roe deer with fallow deer, which the DINOv3-based checkpoint reduced by more than two-thirds, and the labelling of empty frames as animals (848 of the 2,857 empty images, most often cat, fallow deer or sheep for v1.2; Figure A5d), whose number the DINOv3-based checkpoint did not reduce at all because these errors begin with detector false positives rather than with the classifier. Both can be understood as the price of a larger regional vocabulary applied beyond its training sites, where a model with more fine-grained classes has more ways to be wrong, and both suggest specific remedies: more site-diverse deer imagery, hard-negative empty images, of which the current validation split contains none, and the use of sequence-level context~\parencite{dussert2026}.

The value of this vocabulary is greatest where a label triggers an action, and that is also where its errors matter most. In a selective trap or feeder of the kind described in Section 1, the two directions of squirrel confusion carry very different costs: a grey squirrel labelled red is merely missed, whereas a red squirrel labelled grey exposes a protected species to harm. The out-of-distribution results counsel caution here. Although DeepFaune-UK separated the two species almost perfectly on held-out crops from its training sites, the MammalWeb set contained only three red squirrel images, and all three were labelled grey (Figure A5b), while 14 of the 2,742 grey squirrel images were labelled red; DeepFaune-UK v1.3 likewise labelled all three red squirrels grey, although it labelled only 8 grey squirrels red. Three images cannot support an estimate of accuracy, but they are enough to show that in-distribution performance on a rare, safety-critical class does not transfer automatically to other sites. A device that acts on such labels should therefore apply an asymmetric decision rule, acting only when the classifier is highly confident of the target species and defaulting to the safe action otherwise, as the AI-based predator trap described in Section 1 does by arming only when it identifies a target species with high confidence~\parencite{pfnz2025}, and it should be validated with local imagery of the non-target species before deployment. This is, however, an argument for the approach studied here rather than against it. The low adaptation cost observed here makes further local fine-tuning feasible: the model can be re-fine-tuned with images of red squirrels from the sites where a device will operate, an option that is not open to users of a fixed global vocabulary that does not distinguish the two species at all.

The contrast between these two readings also shows why vocabulary-aware evaluation is needed. Scored strictly against exact species names, DeepFaune-UK appears dramatically better than both baselines; scored like-for-like, it is slightly worse than DeepFaune. Both statements are true, and which of them matters depends on whether a user needs species-level UK labels or is content with coarse groups. A single headline accuracy would have concealed one or other of these facts, which is consistent with earlier warnings that reported camera-trap accuracy depends heavily on how and where it is measured~\parencite{beery2018,velez2023}. We therefore recommend that comparisons between camera-trap classifiers with different label sets report results at both levels.

If the benchmark answers RQ1 with a qualified result, the learning curve gives a clearer answer to RQ2, and it is the more consequential finding for the democratisation argument. A classifier trained on 4,270 crops, roughly 150 per class on average, in 22 minutes, with under 4 GB of GPU memory allocated and 0.06 kWh of energy, performed on out-of-distribution images within seed-to-seed variation of one trained on ten times as much data. This contrasts with earlier camera-trap studies using convolutional networks, in which performance improved with the number of training images per class~\parencite{shahinfar2020} and classes with fewer than 500 training images had low and highly variable recall~\parencite{schneider2020}. A drop from the in-distribution test crops to the out-of-distribution images is also visible here (macro-F1 of about 0.99 against about 0.85 like-for-like), although, allowing for the different metrics, it is much smaller than the fall in top-1 accuracy from 95.6\% to 68.7\% reported by Schneider et al.~\parencite{schneider2020}. Accuracy showed the same pattern for the DINOv3-based checkpoint, at roughly twice the training cost (about 46 minutes and 4.4 GB of GPU memory at 10\%), although its like-for-like macro-F1 was about 0.01 lower at 10\% than with all the data. This is consistent with most of the model's ability to generalise coming from the open pretrained weights, and with the role of regional data being to define the label set rather than to teach the model to see. For practitioners, the implication is that a few hundred well-chosen images per species (about 150 crops per class on average sufficed here, for the 12 species with more than three OOD images) may be enough to obtain a useful regional model, and that the marginal value of further labelling lies in covering new sites, conditions and confusable species rather than in volume for its own sake. It also bears on the concerns about the financial and environmental cost of model development raised in the wider literature~\parencite{strubell2019,schwartz2020}, since the cost of adaptation reported here is small compared with the cost of producing the underlying models.

In the terms proposed by Seger et al.~\parencite{seger2023}, this is evidence for the democratisation of AI \emph{development}: a small team without large-scale compute was able to build, and release in a deployable form, a model tailored to regional conservation needs. It is, however, a conditional form of democratisation. It depends entirely on others having trained and released the foundation backbone, the base classifier and the detector, at costs far beyond those reported here, and it depends on the terms under which those artefacts were released. The CC BY-SA licence of the DeepFaune weights requires derived models, if shared, to be shared on the same terms, and the AGPL-3.0 licence of MegaDetector v6 places obligations on software that deploys it. Small actors thus gain access to development, but not to the means of producing foundation models, and they remain dependent on large actors' continued willingness to publish open weights on usable terms. The v1.3 experiment makes this dependence concrete. On the main evaluation set, the largest improvement in like-for-like accuracy among the factors varied in this study came not from the regional team's data or effort but from adopting a newer base model trained and released by others: with no new data and an unchanged recipe, moving to the DINOv3-based classifier closed most of the gap to DeepFaune v1.5.0. Small teams can therefore follow progress at the frontier quickly and cheaply, but they are unlikely to lead it. The same step also shows that the terms of openness can change between releases. DINOv2 is distributed under the permissive Apache 2.0 licence (after an initial non-commercial release), whereas DINOv3 is distributed under Meta's own DINOv3 License, which restricts certain uses and requires derived models to be distributed under its terms~\parencite{dinov3license}, so DeepFaune-UK v1.3 inherits conditions that v1.2 does not. Because the ShareAlike terms of the DeepFaune weights forbid additional restrictions, which the DINOv3 License imposes, the two licences are not obviously compatible, a tension that DeepFaune v1.5.0 itself already carries.

For these reasons we see the roles of large and small actors as complementary rather than competitive. Large initiatives are best placed to provide broad, robust models and backbones, while regional teams can contribute local vocabulary, local test data and independent evaluation, as in this study, where a hand-verified out-of-distribution set exposed strengths and weaknesses in all of the models compared. Small groups cannot match the scale of global models, but with open weights they can still play a substantive part, and a model built in this way can be placed directly in the hands of practitioners through openly available, non-commercial platforms such as Trap Tracker~\parencite{traptracker2026}.

\subsection{Limitations}

Several limitations qualify these findings. The most immediate concerns the out-of-distribution set itself. It comes from a single citizen-science project and is dominated by roe deer, rabbit and grey squirrel, while two species are represented by only three images each; per-species estimates for those species are therefore unreliable, and macro-averaged scores carry correspondingly wide intervals. A larger or more balanced benchmark was not available to us: to our knowledge, no publicly released, independently verified camera-trap dataset from UK sites covers these species in balanced numbers. The set also proved not to be fully independent of our training archive: 1,039 of its images were exact copies of master images, from sequences that we removed, and the remaining images of the overlapping contributors may still come from cameras that contributed training images; only the smaller site-disjoint subset, defined by contributor, is free of known overlap, and because only exact copies were detected even that subset may not be entirely independent. Visual checks also showed that some images labelled empty contain animals the reference labels missed, which counts against every model that reports them. The set's 14 species are also only half of DeepFaune-UK's 28-class vocabulary, so the benchmark does not test the classes, such as capercaillie, curlew and pine marten, that motivated building a regional model in the first place, yet it still penalises DeepFaune-UK whenever it predicts one of them.

The evaluation protocol introduces further caveats. The like-for-like mapping necessarily involves judgement, particularly for SpeciesNet's higher-rank labels, and a different but equally defensible mapping could shift the like-for-like figures slightly. All models were assessed at the image level without sequence information, even though DeepFaune aggregates predictions over sequences by default and SpeciesNet outputs can be smoothed by sequence with the MegaDetector post-processing tools, and the baselines were run at fixed operating points (Section 3.4): DeepFaune-UK's thresholds were checked on validation data, but those of the baselines were not tuned. These choices keep the comparison simple and reproducible, but none of them shows any model at its best.

The resource analysis is similarly bounded. The learning curve used a single training recipe and detector with three seeds, and the DINOv3 backbone was trained at only two of the four data fractions. The resource measurements were taken on one workstation GPU and cover GPU energy only, excluding the host computer and, more importantly, the cost of producing the pretrained weights on which the whole exercise depends. Finally, all DeepFaune-UK results are reported by the model's own developers and have not been independently replicated, although both baselines were run with their developers' official code (Section 3.4), and the DeepFaune-UK v1.3 analysis was added after the v1.2 benchmark had been run on the same OOD set, so although the OOD set played no part in selecting v1.3 or its thresholds, its results on that set are not fully prospective. Likewise, MegaDetector v6 was preferred to a fine-tuned YOLOv8s detector on the basis of its detection recall on this set, and earlier candidate systems were also compared on it (Section 3.3), which may favour the DeepFaune-UK pipeline to an unknown degree; the baselines' detectors were not chosen in this way. The comparison is also necessarily a snapshot: DeepFaune and SpeciesNet are under active development, and future releases may change the picture.

\subsection{Future Work}

These limitations point directly to the next steps. The two dominant error types suggest that adding hard-negative empty images and more site-diverse deer imagery to the training data is the most promising route to closing any remaining like-for-like gap to DeepFaune v1.5.0 on the main set, and whether it does so is an empirical question worth answering; because the empty-frame errors arise in the detector, they may require hard negatives for the detector stage or an explicit empty class in the classifier rather than further classifier training alone. On the resource side, extending the learning curve below 10\% of the training data would locate the point at which performance begins to degrade, and repeating the resource measurements on consumer GPUs would test the claim that this kind of development is within reach of small teams more directly. The evaluation itself should also be broadened to additional projects, regions and seasons, and in particular to the rarer classes that motivated the regional vocabulary but are absent or barely represented in the present out-of-distribution set; red squirrel imagery from regions where the species persists is the most pressing gap, and a field trial of a species-selective device would test whether the model's labels are reliable enough to act on. Finally, evaluating sequence-level aggregation for all of the models, and measuring live performance once DeepFaune-UK is deployed through Trap Tracker, would show how the image-level differences reported here translate into the decisions practitioners actually make.

\section{Conclusion}

This study set out to ask whether openly released model weights allow a small team, working with a modest dataset and a single GPU, to take a meaningful part in AI development for conservation. Using UK wildlife as a case study, we fine-tuned the open DeepFaune classifier on a regional vocabulary, paired it with the unmodified MegaDetector v6 detector, and compared the resulting model with DeepFaune v1.5.0 and SpeciesNet on hand-verified citizen-science images, after removing every sequence that overlapped with our training archive. The regional model achieved a strict image-level accuracy of 89\% on these out-of-distribution images, counting correctly identified empty frames, whereas the general-purpose models, despite their far greater scale, achieved 57--59\%, and it remained within a few points of the best continental model on the coarse groups that both could name. Where the vocabularies overlap, the larger model was nonetheless slightly more accurate on the main evaluation set, though not on a smaller site-disjoint subset, and we have reported that result plainly rather than letting the strict species-level comparison stand alone. Applying the same recipe to the newer DINOv3-based DeepFaune classifier, DeepFaune-UK v1.3 reached a strict accuracy of 91\% and narrowed the like-for-like gap to under one percentage point: changing the starting checkpoint under the same regional training recipe removed most of the gap observed on the main set.

The resource findings carry the broader argument. Comparable out-of-distribution accuracy was obtained from only 10\% of the training data, in about 22 minutes of training on a single GPU with under 4 GB of allocated GPU memory (under an hour and 4.4 GB with the DINOv3 backbone, though with slightly lower macro-F1), which is consistent with the generalisation of such a model deriving largely from the open pretrained weights and with regional data serving mainly to define what the model should name. Taken together, these results suggest that open foundation weights make meaningful AI development feasible for small, domain-led teams, but that this democratisation is conditional: it depends on well-resourced developers continuing to train and release such weights, and on the licences under which they choose to do so, which, as the move from DINOv2 to DINOv3 shows, can change from one release to the next. Small groups cannot compete with the largest players at scale, but by building on openly shared models, contributing local vocabulary and independent evaluation, and returning their models to practitioners through open platforms, they can still play a substantive part in the development of AI for conservation.

\section*{Author Contributions}

Conceptualization, methodology, software, validation, formal analysis, investigation, data curation, writing --- original draft preparation, and writing --- review and editing: P.F. The author has read and agreed to the submitted version of the manuscript.

\section*{Funding}

This research received no external funding.

\section*{Institutional Review Board Statement}

Not applicable. This study involved no direct handling, capture, or manipulation of live animals: all imagery was collected non-invasively by static, motion-triggered camera traps, either at contributing sites under prior data-sharing and landowner-access agreements (master dataset) or by MammalWeb citizen scientists and obtained from its openly licensed dataset (out-of-distribution set); no additional data collection was undertaken for this study.

\section*{Informed Consent Statement}

Not applicable, as this study did not involve human subjects.

\section*{Data Availability Statement}

The master dataset contains camera-trap data shared under prior data-sharing agreements with contributing sites and is not publicly available in full. The out-of-distribution set is derived from the MammalWeb gold-standard dataset, which is openly available under a CC BY-SA 4.0 licence~\parencite{hsing2018data}. The DeepFaune~\parencite{deepfaune2026}, MegaDetector~\parencite{hernandez2024} and SpeciesNet~\parencite{google2025} models are openly available from their respective sources. Code for the benchmark, learning-curve and scoring analyses is available from the author on reasonable request. DeepFaune-UK v1.2 is openly available on Zenodo as ONNX models~\parencite{fergus2026v12}, with the classifier released under CC BY-SA 4.0, as required by the ShareAlike terms of the DeepFaune weights, and the detector, an unmodified export of MegaDetector v6, under its original AGPL-3.0 licence. DeepFaune-UK v1.3 is available on Zenodo in the same format~\parencite{fergus2026v13}; its detector is released under AGPL-3.0, and its classifier, being derived from DeepFaune v1.5.0 and from DINOv3 weights, is distributed under the terms of the DINOv3 License~\parencite{dinov3license} as well as the attribution and ShareAlike terms of the DeepFaune weights.

\section*{Acknowledgments}

The author thanks the DeepFaune initiative, the MegaDetector and PyTorch-Wildlife team, and Google's SpeciesNet team for openly releasing the models on which this work depends; the MammalWeb project and its citizen scientists for the out-of-distribution imagery; the Game \& Wildlife Conservation Trust, the British Hedgehog Preservation Society and the People's Trust for Endangered Species for the collaborations this programme builds on; and the Conservation AI and Trap Tracker development team for the underlying infrastructure.

\section*{Conflicts of Interest}

The author declares no financial conflict of interest. Trap Tracker and the Conservation AI platform referenced in this paper are non-commercial research infrastructure developed by the author. The baseline models benchmarked here are third-party systems, run with their developers' official code (Section 3.4).

\printbibliography

@report{mathews2018,
  author      = {Mathews, F. and Kubasiewicz, L. M. and Gurnell, J. and Harrower, C. A. and McDonald, R. A. and Shore, R. F.},
  title       = {A Review of the Population and Conservation Status of {British} Mammals},
  institution = {Natural England},
  location    = {Peterborough, UK},
  year        = {2018},
  number      = {JP025},
  type        = {Report}
}

@report{burns2023,
  author      = {Burns, F. and Mordue, S. and al Fulaij, N. and Boersch-Supan, P. H. and Boswell, J. and Boyd, R. J. and Bradfer-Lawrence, T. and de Ornellas, P. and de Palma, A. and de Zylva, P. and others},
  title       = {State of Nature 2023},
  institution = {State of Nature Partnership},
  year        = {2023},
  type        = {Report},
  pagetotal   = {214}
}

@inproceedings{beery2019,
  author     = {Beery, S. and Morris, D. and Yang, S.},
  title      = {Efficient Pipeline for Camera Trap Image Review},
  booktitle  = {Data Mining and AI for Conservation Workshop at KDD 2019},
  location   = {Anchorage, AK, USA},
  year       = {2019},
  eprint     = {1907.06772},
  eprinttype = {arXiv}
}

@online{hernandez2024,
  author     = {Hernandez, A. and Miao, Z. and Vargas, L. and Beery, S. and Dodhia, R. and Arbelaez, P. and Lavista Ferres, J. M.},
  title      = {{Pytorch-Wildlife}: A Collaborative Deep Learning Framework for Conservation},
  year       = {2024},
  eprint     = {2405.12930},
  eprinttype = {arXiv}
}

@article{rigoudy2023,
  author  = {Rigoudy, N. and Dussert, G. and Benyoub, A. and others},
  title   = {The {DeepFaune} Initiative: A Collaborative Effort towards the Automatic Identification of {European} Fauna in Camera Trap Images},
  journal = {European Journal of Wildlife Research},
  number  = {6},
  year    = {2023},
  volume  = {69},
  eid     = {113},
  doi     = {10.1007/s10344-023-01742-7}
}

@article{gadot2024,
  author  = {Gadot, T. and Istrate, {	extcommabelow{S}}. and Kim, H. and Morris, D. and Beery, S. and Birch, T. and Ahumada, J.},
  title   = {To Crop or Not to Crop: Comparing Whole-Image and Cropped Classification on a Large Dataset of Camera Trap Images},
  journal = {IET Computer Vision},
  number  = {8},
  year    = {2024},
  volume  = {18},
  pages   = {1193--1208},
  doi     = {10.1049/cvi2.12318}
}

@software{google2025,
  author       = {{Google}},
  title        = {{SpeciesNet}: An Ensemble of {AI} Models for Classifying Wildlife in Camera Trap Images},
  howpublished = {GitHub repository},
  year         = {2025},
  url          = {https://github.com/google/cameratrapai},
  urldate      = {2026-10-09}
}

@article{velez2023,
  author  = {V{\'e}lez, J. and McShea, W. and Shamon, H. and Castiblanco-Camacho, P. J. and Tabak, M. A. and Chalmers, C. and Fergus, P. and Fieberg, J.},
  title   = {An Evaluation of Platforms for Processing Camera-Trap Data Using Artificial Intelligence},
  journal = {Methods in Ecology and Evolution},
  number  = {2},
  year    = {2023},
  volume  = {14},
  pages   = {459--477},
  doi     = {10.1111/2041-210X.14044}
}

@online{ahmed2020,
  author     = {Ahmed, N. and Wahed, M.},
  title      = {The De-Democratization of {AI}: Deep Learning and the Compute Divide in Artificial Intelligence Research},
  year       = {2020},
  eprint     = {2010.15581},
  eprinttype = {arXiv}
}

@inproceedings{seger2023,
  author    = {Seger, E. and Ovadya, A. and Siddarth, D. and Garfinkel, B. and Dafoe, A.},
  title     = {Democratising {AI}: Multiple Meanings, Goals, and Methods},
  booktitle = {Proceedings of the 2023 AAAI/ACM Conference on AI, Ethics, and Society (AIES '23)},
  location  = {Montr{\'e}al, QC, Canada},
  year      = {2023},
  pages     = {715--722},
  doi       = {10.1145/3600211.3604693}
}

@article{chalmers2025,
  author  = {Chalmers, C. and Fergus, P. and Wich, S. and Longmore, S. N. and Walsh, N. D. and Oliver, L. and Warrington, J. and Quinlan, J. and Appleby, K.},
  title   = {{AI}-Driven Real-Time Monitoring of Ground-Nesting Birds: A Case Study on Curlew Detection Using {YOLOv10}},
  journal = {Remote Sensing},
  number  = {5},
  year    = {2025},
  volume  = {17},
  eid     = {769},
  doi     = {10.3390/rs17050769}
}

@online{fergus2026,
  author     = {Fergus, P. and Stephens, P. and Hill, R. A. and Oliver, L. and Appleby, K. and Beatham, S. and Walsh, N. D. and Nixon, S. and Matthews, N. and Sutherland, C. and Hitchcock, K.},
  title      = {Democratising Camera Trap {AI}: An Open-Source Model for Detecting {UK} Mammals},
  year       = {2026},
  eprint     = {2606.10940},
  eprinttype = {arXiv}
}

@software{traptracker2026,
  author       = {{Trap Tracker}},
  title        = {Trap Tracker Real-Time Camera Application},
  howpublished = {GitHub repository},
  year         = {2026},
  url          = {https://github.com/traptracker-ai/traptracker-realtime-camera},
  urldate      = {2026-09-25}
}

@article{hsing2018,
  author  = {Hsing, P.-Y. and Bradley, S. and Kent, V. T. and Hill, R. A. and Smith, G. C. and Whittingham, M. J. and Cokill, J. and Crawley, D. and {MammalWeb volunteers} and Stephens, P. A.},
  title   = {Economical Crowdsourcing for Camera Trap Image Classification},
  journal = {Remote Sensing in Ecology and Conservation},
  number  = {4},
  year    = {2018},
  volume  = {4},
  pages   = {361--374},
  doi     = {10.1002/rse2.84}
}

@article{norouzzadeh2018,
  author  = {Norouzzadeh, M. S. and Nguyen, A. and Kosmala, M. and Swanson, A. and Palmer, M. S. and Packer, C. and Clune, J.},
  title   = {Automatically Identifying, Counting, and Describing Wild Animals in Camera-Trap Images with Deep Learning},
  journal = {Proceedings of the National Academy of Sciences},
  number  = {25},
  year    = {2018},
  volume  = {115},
  pages   = {E5716--E5725},
  doi     = {10.1073/pnas.1719367115}
}

@article{willi2019,
  author  = {Willi, M. and Pitman, R. T. and Cardoso, A. W. and Locke, C. and Swanson, A. and Boyer, A. and Veldthuis, M. and Fortson, L.},
  title   = {Identifying Animal Species in Camera Trap Images Using Deep Learning and Citizen Science},
  journal = {Methods in Ecology and Evolution},
  number  = {1},
  year    = {2019},
  volume  = {10},
  pages   = {80--91},
  doi     = {10.1111/2041-210X.13099}
}

@inproceedings{beery2018,
  author    = {Beery, S. and Van Horn, G. and Perona, P.},
  title     = {Recognition in Terra Incognita},
  booktitle = {Computer Vision -- ECCV 2018},
  series    = {Lecture Notes in Computer Science},
  publisher = {Springer},
  location  = {Cham, Switzerland},
  year      = {2018},
  pages     = {472--489},
  doi       = {10.1007/978-3-030-01270-0_28}
}

@article{oquab2024,
  author     = {Oquab, M. and Darcet, T. and Moutakanni, T. and Vo, H. and Szafraniec, M. and Khalidov, V. and Fernandez, P. and Haziza, D. and Massa, F. and El-Nouby, A. and others},
  title      = {{DINOv2}: Learning Robust Visual Features without Supervision},
  journal    = {Transactions on Machine Learning Research},
  year       = {2024},
  eprint     = {2304.07193},
  eprinttype = {arXiv}
}

@online{simeoni2025,
  author     = {Sim{\'e}oni, O. and Vo, H. V. and Seitzer, M. and Baldassarre, F. and Oquab, M. and Jose, C. and Khalidov, V. and Szafraniec, M. and Yi, S. and Ramamonjisoa, M. and others},
  title      = {{DINOv3}},
  year       = {2025},
  eprint     = {2508.10104},
  eprinttype = {arXiv}
}

@inproceedings{strubell2019,
  author    = {Strubell, E. and Ganesh, A. and McCallum, A.},
  title     = {Energy and Policy Considerations for Deep Learning in {NLP}},
  booktitle = {Proceedings of the 57th Annual Meeting of the Association for Computational Linguistics (ACL)},
  location  = {Florence, Italy},
  year      = {2019},
  pages     = {3645--3650},
  doi       = {10.18653/v1/P19-1355}
}

@article{schwartz2020,
  author  = {Schwartz, R. and Dodge, J. and Smith, N. A. and Etzioni, O.},
  title   = {Green {AI}},
  journal = {Communications of the ACM},
  number  = {12},
  year    = {2020},
  volume  = {63},
  pages   = {54--63},
  doi     = {10.1145/3381831}
}

@article{tuia2022,
  author  = {Tuia, D. and Kellenberger, B. and Beery, S. and Costelloe, B. R. and Zuffi, S. and Risse, B. and Mathis, A. and Mathis, M. W. and van Langevelde, F. and Burghardt, T. and others},
  title   = {Perspectives in Machine Learning for Wildlife Conservation},
  journal = {Nature Communications},
  year    = {2022},
  volume  = {13},
  pages   = {792},
  doi     = {10.1038/s41467-022-27980-y}
}

@online{chalmers2019,
  author     = {Chalmers, C. and Fergus, P. and Wich, S. and Curbelo Monta{\~n}ez, C. A.},
  title      = {Conservation {AI}: Live Stream Analysis for the Detection of Endangered Species Using Convolutional Neural Networks and Drone Technology},
  year       = {2019},
  eprint     = {1910.07360},
  eprinttype = {arXiv}
}

@online{ljmu2024,
  author       = {{Liverpool John Moores University}},
  title        = {{AI} to Help National Hedgehog Survey},
  organization = {LJMU News},
  date         = {2024-03-25},
  url          = {https://www.ljmu.ac.uk/about-us/news/articles/2024/3/25/hedgehogs},
  urldate      = {2026-09-25}
}

@article{fergus2024,
  author  = {Fergus, P. and Chalmers, C. and Matthews, N. and Nixon, S. and Burger, A. and Hartley, O. and Sutherland, C. and Lambin, X. and Longmore, S. and Wich, S.},
  title   = {Towards Context-Rich Automated Biodiversity Assessments: Deriving {AI}-Powered Insights from Camera Trap Data},
  journal = {Sensors},
  number  = {24},
  year    = {2024},
  volume  = {24},
  eid     = {8122},
  doi     = {10.3390/s24248122}
}

@inproceedings{recht2019,
  author    = {Recht, B. and Roelofs, R. and Schmidt, L. and Shankar, V.},
  title     = {Do {ImageNet} Classifiers Generalize to {ImageNet}?},
  booktitle = {Proceedings of the 36th International Conference on Machine Learning (ICML)},
  location  = {Long Beach, CA, USA},
  year      = {2019},
  pages     = {5389--5400}
}

@article{barz2020,
  author  = {Barz, B. and Denzler, J.},
  title   = {Do We Train on Test Data? Purging {CIFAR} of Near-Duplicates},
  journal = {Journal of Imaging},
  number  = {6},
  year    = {2020},
  volume  = {6},
  pages   = {41},
  doi     = {10.3390/jimaging6060041}
}

@article{kapoor2023,
  author  = {Kapoor, S. and Narayanan, A.},
  title   = {Leakage and the Reproducibility Crisis in Machine-Learning-Based Science},
  journal = {Patterns},
  number  = {9},
  year    = {2023},
  volume  = {4},
  pages   = {100804},
  doi     = {10.1016/j.patter.2023.100804}
}

@inproceedings{yosinski2014,
  author    = {Yosinski, J. and Clune, J. and Bengio, Y. and Lipson, H.},
  title     = {How Transferable Are Features in Deep Neural Networks?},
  booktitle = {Advances in Neural Information Processing Systems 27 (NIPS 2014)},
  location  = {Montreal, Canada},
  year      = {2014},
  pages     = {3320--3328}
}

@inproceedings{howard2018,
  author    = {Howard, J. and Ruder, S.},
  title     = {Universal Language Model Fine-Tuning for Text Classification},
  booktitle = {Proceedings of the 56th Annual Meeting of the Association for Computational Linguistics (ACL)},
  location  = {Melbourne, Australia},
  year      = {2018},
  pages     = {328--339}
}

@inproceedings{loshchilov2019,
  author     = {Loshchilov, Ilya and Hutter, Frank},
  title      = {Decoupled Weight Decay Regularization},
  booktitle  = {International Conference on Learning Representations},
  year       = {2019},
  eprint     = {1711.05101},
  eprinttype = {arXiv}
}

@article{buda2018,
  author  = {Buda, M. and Maki, A. and Mazurowski, M. A.},
  title   = {A Systematic Study of the Class Imbalance Problem in Convolutional Neural Networks},
  journal = {Neural Networks},
  year    = {2018},
  volume  = {106},
  pages   = {249--259},
  doi     = {10.1016/j.neunet.2018.07.011}
}

@inproceedings{szegedy2016,
  author    = {Szegedy, C. and Vanhoucke, V. and Ioffe, S. and Shlens, J. and Wojna, Z.},
  title     = {Rethinking the Inception Architecture for Computer Vision},
  booktitle = {Proceedings of the IEEE Conference on Computer Vision and Pattern Recognition (CVPR)},
  location  = {Las Vegas, NV, USA},
  year      = {2016},
  pages     = {2818--2826}
}

@inproceedings{muller2019,
  author    = {M{\"u}ller, R. and Kornblith, S. and Hinton, G.},
  title     = {When Does Label Smoothing Help?},
  booktitle = {Proceedings of the 33rd Conference on Neural Information Processing Systems (NeurIPS)},
  location  = {Vancouver, Canada},
  year      = {2019},
  pages     = {4694--4703}
}

@book{efron1993,
  title={An introduction to the bootstrap},
  author={Efron, Bradley and Tibshirani, Robert J},
  year={1994},
  publisher={Chapman and Hall/CRC}
}

@article{mcnemar1947,
  author  = {McNemar, Q.},
  title   = {Note on the Sampling Error of the Difference between Correlated Proportions or Percentages},
  journal = {Psychometrika},
  number  = {2},
  year    = {1947},
  volume  = {12},
  pages   = {153--157},
  doi     = {10.1007/BF02295996}
}

@article{dussert2026,
  author  = {Dussert, G. and Dray, S. and Chamaill{\'e}-Jammes, S. and Miele, V.},
  title   = {Paying Attention to Other Animal Detections Improves Camera Trap Classification Models},
  journal = {Methods in Ecology and Evolution},
  number  = {4},
  year    = {2026},
  volume  = {17},
  pages   = {1248--1258},
  doi     = {10.1111/2041-210x.70260}
}

@report{desnz2024,
  author      = {{Department for Energy Security and Net Zero}},
  title       = {Greenhouse Gas Reporting: Conversion Factors 2024},
  institution = {DESNZ},
  location    = {London, UK},
  type        = {Dataset},
  year        = {2024},
  url         = {https://www.gov.uk/government/publications/greenhouse-gas-reporting-conversion-factors-2024},
  urldate     = {2026-10-09}
}

@article{rushton2006,
  author  = {Rushton, S. P. and Lurz, P. W. W. and Gurnell, J. and Nettleton, P. and Bruemmer, C. and Shirley, M. D. F. and Sainsbury, A. W.},
  title   = {Disease Threats Posed by Alien Species: The Role of a Poxvirus in the Decline of the Native Red Squirrel in {Britain}},
  journal = {Epidemiology and Infection},
  number  = {3},
  year    = {2006},
  volume  = {134},
  pages   = {521--533},
  doi     = {10.1017/S0950268805005303}
}

@legislation{wca1981,
  title   = {{Wildlife and Countryside Act 1981}, c. 69, Schedule 5},
  year    = {1981},
  url     = {https://www.legislation.gov.uk/ukpga/1981/69/schedule/5},
  urldate = {2026-10-12}
}

@legislation{ias2019,
  title   = {{The Invasive Alien Species (Enforcement and Permitting) Order 2019}, {SI} 2019/527},
  year    = {2019},
  url     = {https://www.legislation.gov.uk/uksi/2019/527},
  urldate = {2026-10-12}
}

@article{beatham2023,
  author  = {Beatham, S. E. and Coats, J. and Stephens, P. A. and Massei, G.},
  title   = {Factors Affecting Bait Uptake by the Grey Squirrel (\emph{Sciurus carolinensis}) and the Future Delivery of Oral Contraceptives},
  journal = {Wildlife Research},
  eid     = {WR22159},
  year    = {2024},
  volume  = {51},
  number  = {1},
  doi     = {10.1071/WR22159}
}

@article{read2019,
  author  = {Read, J. L. and Bowden, T. and Hodgens, P. and Hess, M. and McGregor, H. and Moseby, K.},
  title   = {Target Specificity of the {Felixer} Grooming ``Trap''},
  journal = {Wildlife Society Bulletin},
  number  = {1},
  year    = {2019},
  volume  = {43},
  pages   = {112--120},
  doi     = {10.1002/wsb.942}
}

@online{pfnz2025,
  author  = {{Predator Free NZ Trust}},
  title   = {Next-Gen Traps Built for {NZ}'s Toughest Terrain},
  year    = {2025},
  url     = {https://predatorfreenz.org/research/next-gen-traps-built-for-nz-toughest-terrain/},
  urldate = {2026-10-12}
}

@article{sheehy2014,
  author  = {Sheehy, E. and Lawton, C.},
  title   = {Population Crash in an Invasive Species following the Recovery of a Native Predator: The Case of the {American} Grey Squirrel and the {European} Pine Marten in {Ireland}},
  journal = {Biodiversity and Conservation},
  number  = {3},
  year    = {2014},
  volume  = {23},
  pages   = {753--774},
  doi     = {10.1007/s10531-014-0632-7}
}

@article{sheehy2018,
  author  = {Sheehy, E. and Sutherland, C. and O'Reilly, C. and Lambin, X.},
  title   = {The Enemy of My Enemy Is My Friend: Native Pine Marten Recovery Reverses the Decline of the Red Squirrel by Suppressing Grey Squirrel Populations},
  journal = {Proceedings of the Royal Society B: Biological Sciences},
  number  = {1874},
  year    = {2018},
  volume  = {285},
  pages   = {20172603},
  doi     = {10.1098/rspb.2017.2603}
}

@legislation{deeract1991,
  title   = {{Deer Act 1991}, c. 54, Schedule 1},
  year    = {1991},
  url     = {https://www.legislation.gov.uk/ukpga/1991/54/schedule/1},
  urldate = {2026-10-12}
}

@article{barton2026,
  author  = {Barton, M. G. and Conway, G. J. and Henderson, I. G. and Baddams, J. and Balchin, C. S. and Brides, K. and Butcher, N. and Cameron, T. C. and Davis, T. and Eyre, J. and others},
  title   = {Meta-Analysis of Predator Identity in Nest-Camera Studies in the {British Islands}},
  journal = {Ibis},
  year    = {2026},
  volume  = {168},
  pages   = {42--62},
  doi     = {10.1111/ibi.13436}
}

@article{kitchener2005,
  author  = {Kitchener, A. C. and Yamaguchi, N. and Ward, J. M. and Macdonald, D. W.},
  title   = {A Diagnosis for the {Scottish} Wildcat (\emph{Felis silvestris}): A Tool for Conservation Action for a Critically-Endangered Felid},
  journal = {Animal Conservation},
  number  = {3},
  year    = {2005},
  volume  = {8},
  pages   = {223--237},
  doi     = {10.1017/S1367943005002301}
}

@article{barreto1998,
  author  = {Barreto, G. R. and Rushton, S. P. and Strachan, R. and Macdonald, D. W.},
  title   = {The Role of Habitat and Mink Predation in Determining the Status and Distribution of Water Voles in {England}},
  journal = {Animal Conservation},
  number  = {2},
  year    = {1998},
  volume  = {1},
  pages   = {129--137},
  doi     = {10.1111/j.1469-1795.1998.tb00020.x}
}

@online{dinov3license,
  author  = {{Meta Platforms, Inc.}},
  title   = {{DINOv3} License},
  year    = {2025},
  url     = {https://github.com/facebookresearch/dinov3/blob/main/LICENSE.md},
  urldate = {2026-10-13}
}

@dataset{fergus2026v12,
  author    = {Fergus, P.},
  title     = {{DeepFaune-UK} v1.2: {ONNX} Model Weights for {UK} Mammal and Bird Species Detection and Classification},
  publisher = {Zenodo},
  version   = {1.2},
  year      = {2026},
  url       = {https://zenodo.org/records/23314735}
}

@dataset{fergus2026v13,
  author    = {Fergus, P.},
  title     = {{DeepFaune-UK} v1.3: {ONNX} Model Weights for {UK} Mammal and Bird Species Detection and Classification},
  publisher = {Zenodo},
  version   = {1.3},
  year      = {2026},
  url       = {https://zenodo.org/records/23343143}
}

@article{schneider2020,
  author  = {Schneider, S. and Greenberg, S. and Taylor, G. W. and Kremer, S. C.},
  title   = {Three Critical Factors Affecting Automated Image Species Recognition Performance for Camera Traps},
  journal = {Ecology and Evolution},
  number  = {7},
  year    = {2020},
  volume  = {10},
  pages   = {3503--3517},
  doi     = {10.1002/ece3.6147}
}

@article{shahinfar2020,
  author  = {Shahinfar, S. and Meek, P. and Falzon, G.},
  title   = {``How Many Images Do {I} Need?'' Understanding How Sample Size per Class Affects Deep Learning Model Performance Metrics for Balanced Designs in Autonomous Wildlife Monitoring},
  journal = {Ecological Informatics},
  year    = {2020},
  volume  = {57},
  pages   = {101085},
  doi     = {10.1016/j.ecoinf.2020.101085}
}

@article{tabak2020,
  author  = {Tabak, M. A. and Norouzzadeh, M. S. and Wolfson, D. W. and Newton, E. J. and Boughton, R. K. and Ivan, J. S. and Odell, E. A. and Newkirk, E. S. and Conrey, R. Y. and Stenglein, J. and others},
  title   = {Improving the Accessibility and Transferability of Machine Learning Algorithms for Identification of Animals in Camera Trap Images: {MLWIC2}},
  journal = {Ecology and Evolution},
  number  = {19},
  year    = {2020},
  volume  = {10},
  pages   = {10374--10383},
  doi     = {10.1002/ece3.6692}
}

@article{norouzzadeh2021,
  author  = {Norouzzadeh, M. S. and Morris, D. and Beery, S. and Joshi, N. and Jojic, N. and Clune, J.},
  title   = {A Deep Active Learning System for Species Identification and Counting in Camera Trap Images},
  journal = {Methods in Ecology and Evolution},
  number  = {1},
  year    = {2021},
  volume  = {12},
  pages   = {150--161},
  doi     = {10.1111/2041-210X.13504}
}

@inproceedings{stevens2024,
  author    = {Stevens, S. and Wu, J. and Thompson, M. J. and Campolongo, E. G. and Song, C. H. and Carlyn, D. E. and Dong, L. and Dahdul, W. M. and Stewart, C. and Berger-Wolf, T. and others},
  title     = {{BioCLIP}: A Vision Foundation Model for the Tree of Life},
  booktitle = {Proceedings of the IEEE/CVF Conference on Computer Vision and Pattern Recognition (CVPR)},
  location  = {Seattle, WA, USA},
  year      = {2024},
  pages     = {19412--19424},
  doi       = {10.1109/CVPR52733.2024.01836}
}

@article{gabeff2024,
  author  = {Gabeff, V. and Ru{\ss}wurm, M. and Tuia, D. and Mathis, A.},
  title   = {{WildCLIP}: Scene and Animal Attribute Retrieval from Camera Trap Data with Domain-Adapted Vision-Language Models},
  journal = {International Journal of Computer Vision},
  number  = {9},
  year    = {2024},
  volume  = {132},
  pages   = {3770--3786},
  doi     = {10.1007/s11263-024-02026-6}
}

@inproceedings{sajun2025,
  author    = {Sajun, A. R. and Gharaibeh, M. and Rahmeh, N. and Zualkernan, I. and Habib, A. and Gardner, A.},
  title     = {Caught in the Lens: Zero-Shot {BioCLIP} Struggles with Camera Trap Wildlife Images},
  booktitle = {Smart Objects and Technologies for Social Good (GOODTECHS 2024)},
  series    = {Lecture Notes of the Institute for Computer Sciences, Social Informatics and Telecommunications Engineering},
  publisher = {Springer},
  location  = {Cham, Switzerland},
  volume  = {648},
  year      = {2025},
  pages     = {131--145},
  doi       = {10.1007/978-3-032-01472-6_11}
}

@online{jeon2026,
  author     = {Jeon, S. and Tian, H. and Wang, L. and Mai, Z. and Bakshi, V. and Hou, J. and Zhang, P. and Chowdhury, A. and Gu, J. and Chao, W.-L.},
  title      = {Lessons and Open Questions from a Unified Study of Camera-Trap Species Recognition Over Time},
  year       = {2026},
  eprint     = {2603.20509},
  eprinttype = {arXiv}
}

@article{widder2024,
  author  = {Widder, D. G. and Whittaker, M. and West, S. M.},
  title   = {Why `Open' {AI} Systems Are Actually Closed, and Why This Matters},
  journal = {Nature},
  number  = {8040},
  year    = {2024},
  volume  = {635},
  pages   = {827--833},
  doi     = {10.1038/s41586-024-08141-1}
}

@inproceedings{liesenfeld2024,
  author    = {Liesenfeld, A. and Dingemanse, M.},
  title     = {Rethinking Open Source Generative {AI}: Open Washing and the {EU AI Act}},
  booktitle = {Proceedings of the 2024 ACM Conference on Fairness, Accountability, and Transparency (FAccT '24)},
  location  = {Rio de Janeiro, Brazil},
  year      = {2024},
  pages     = {1774--1787},
  doi       = {10.1145/3630106.3659005}
}

@article{kapoor2024,
  title={On the societal impact of open foundation models},
  author={Kapoor, Sayash and Bommasani, Rishi and Klyman, Kevin and Longpre, Shayne and Ramaswami, Ashwin and Cihon, Peter and Hopkins, Aspen and Bankston, Kevin and Biderman, Stella and Bogen, Miranda and others},
  journal={arXiv preprint arXiv:2403.07918},
  year={2024}
}

@dataset{hsing2018data,
  author     = {Hsing, P.-Y. and Bradley, S. and Kent, V. T. and Hill, R. A. and Smith, G. C. and Whittingham, M. J. and Cokill, J. and Crawley, D. and {MammalWeb volunteers} and Stephens, P. A.},
  title      = {Economical Crowdsourcing for Camera Trap Image Classification},
  titleaddon = {Data and gold-standard photos},
  publisher  = {OSF},
  year       = {2015},
  doi        = {10.17605/OSF.IO/ZNM6K},
  note       = {CC BY-SA 4.0}
}

@report{uksa2023,
  author      = {{UK Squirrel Accord}},
  title       = {Grey Squirrel Fertility Control Research Update -- {November} 2023},
  institution = {UK Squirrel Accord and Animal and Plant Health Agency},
  type        = {Research update},
  year        = {2023},
  url         = {https://www.charteredforesters.org/wp-content/uploads/2023/12/UKSA-APHA_grey_squirrel_fertility_control_research_update_final_2023.11.docx.pdf},
  urldate     = {2026-10-21}
}

@article{field2007,
  author  = {Field, C. A. and Welsh, A. H.},
  title   = {Bootstrapping Clustered Data},
  journal = {Journal of the Royal Statistical Society Series B: Statistical Methodology},
  year    = {2007},
  volume  = {69},
  number  = {3},
  pages   = {369--390},
  doi     = {10.1111/j.1467-9868.2007.00593.x}
}

@article{edwards1948,
  author  = {Edwards, A. L.},
  title   = {Note on the ``Correction for Continuity'' in Testing the Significance of the Difference between Correlated Proportions},
  journal = {Psychometrika},
  year    = {1948},
  volume  = {13},
  number  = {3},
  pages   = {185--187},
  doi     = {10.1007/BF02289261}
}

@software{deepfaune2026,
  author       = {{DeepFaune initiative}},
  title        = {{DeepFaune} Software},
  version      = {1.5.0},
  year         = {2026},
  url          = {https://plmlab.math.cnrs.fr/deepfaune/software},
  urldate      = {2026-10-22}
}

@software{megadetector5,
  author       = {{MegaDetector team}},
  title        = {{MegaDetector} 5.0},
  year         = {2022},
  url          = {https://github.com/agentmorris/MegaDetector/releases/tag/v5.0},
  urldate      = {2026-10-22}
}

@dataset{pytorchwildlife2025,
  editor       = {Miao, Z. and Hern{\'a}ndez, A. A. and Vargas, L. F. and Chac{\'o}n, I. D. and Ruiz, D. and Chalmers, C.},
  title        = {{PyTorch-Wildlife} Model Weights},
  version      = {v22},
  publisher    = {Zenodo},
  year         = {2025},
  doi          = {10.5281/zenodo.15398270}
}

@software{traptrackeroffline2026,
  author       = {{Trap Tracker}},
  title        = {Trap Tracker Offline Processor},
  howpublished = {GitHub repository},
  year         = {2026},
  url          = {https://github.com/traptracker-ai/traptracker-offline-processor},
  urldate      = {2026-10-22}
}

\appendix
\section{Example Images}
\setcounter{figure}{0}
\renewcommand{\thefigure}{A\arabic{figure}}
\renewcommand{\theHfigure}{A\arabic{figure}}
\setcounter{table}{0}
\renewcommand{\thetable}{A\arabic{table}}
\renewcommand{\theHtable}{A\arabic{table}}

Figures A1--A5 show example images from the OOD set together with the image-level label produced by each model. They were chosen by hand for legibility from pools selected automatically from the per-image results (for example, images on which both DeepFaune-UK models named the exact species while neither baseline returned the correct harmonised group), so they illustrate the patterns quantified in Section 4 rather than estimating their frequency. Tiles containing an animal are cropped around the annotated animal for display; all models processed the full frame. All images are from the MammalWeb gold-standard dataset~\parencite{hsing2018data} and are reproduced under the Creative Commons Attribution-ShareAlike 4.0 licence with attribution to MammalWeb; they have been cropped and annotated for display. Labels are marked \ding{51} when they name the exact species (or correctly report an empty frame), $\approx$ when they are correct only at the harmonised group level, and \ding{55} when they match neither the species nor its harmonised group, which includes generic labels such as ``bird'', ``mammal'' or ``animal''.

Figure A1 illustrates species-level resolution: both DeepFaune-UK models name the species, whereas DeepFaune v1.5.0 can return only the coarser group its vocabulary contains (squirrel, lagomorph, columbiform or corvid bird) and SpeciesNet sometimes rolls up to a family or a generic ``bird'' or ``animal''. Figures A2 and A3 show difficult UK images, including partial views, night-time infrared frames and motion blur, on which neither baseline returned the correct group: SpeciesNet usually backed off to a label such as ``bird'', ``mammal'' or ``cervidae family'', which is correct but too coarse, and DeepFaune v1.5.0 in two cases named a species absent from Britain (raccoon and chamois).

\begin{figure}[htbp]
\centering
\includegraphics[width=\textwidth]{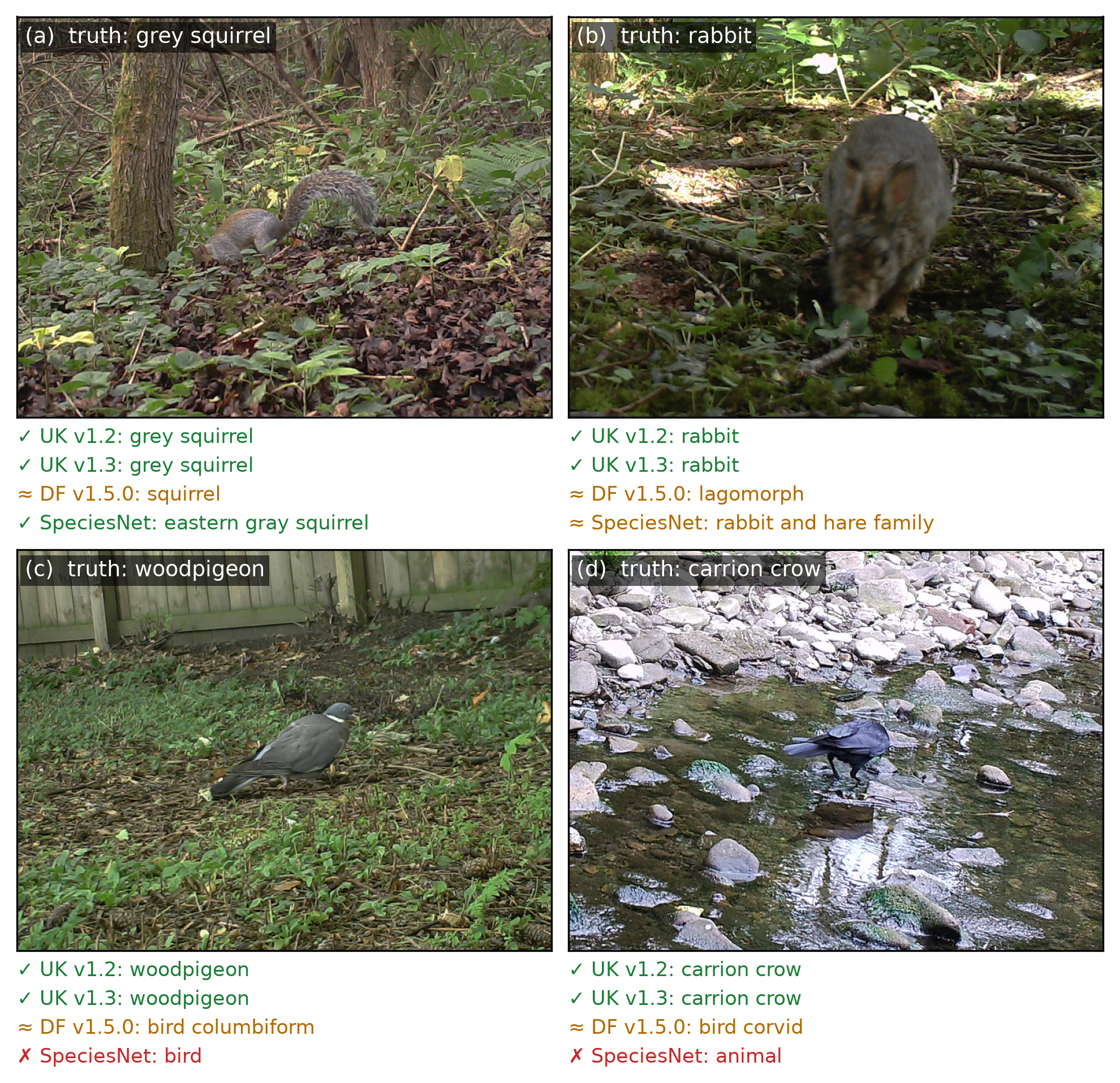}
\caption{Species-level resolution. DeepFaune-UK v1.2 and v1.3 name the species, while DeepFaune v1.5.0 names only a coarser group: (a) grey squirrel, (b) rabbit, (c) woodpigeon and (d) carrion crow. \ding{51} exact species; $\approx$ correct group only; \ding{55} neither (wrong or too coarse). Tiles are cropped around the annotated animal for display. Images: MammalWeb, CC BY-SA 4.0.}
\label{fig:A1}
\end{figure}

\begin{figure}[htbp]
\centering
\includegraphics[width=\textwidth]{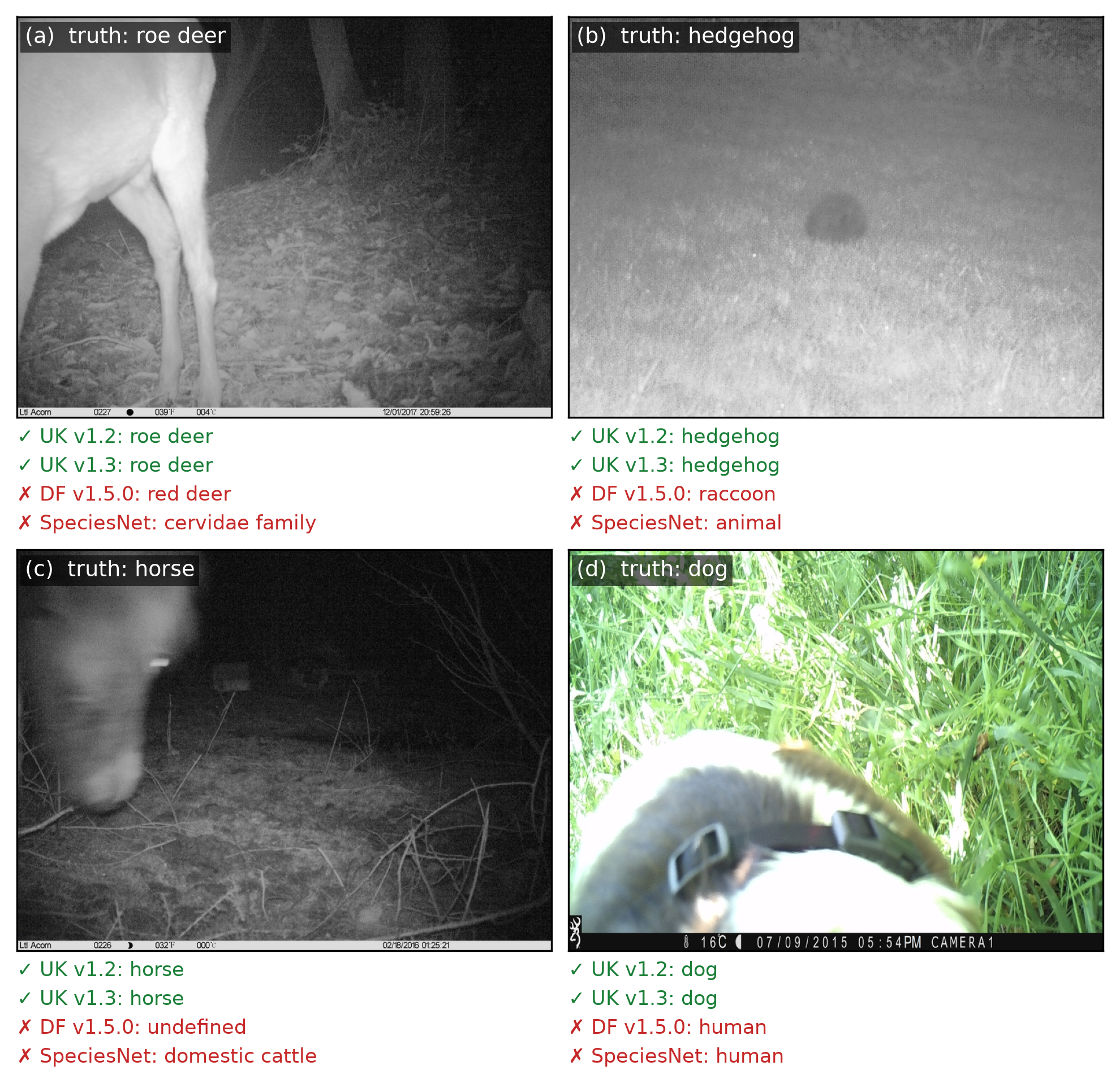}
\caption{Images on which neither baseline returns the correct harmonised group (some baseline labels are correct but too coarse, such as ``cervidae family'' or ``animal'') while both DeepFaune-UK models name the exact species: (a) roe deer, (b) hedgehog, (c) horse and (d) dog. Symbols as in Figure A1.}
\label{fig:A2}
\end{figure}

\begin{figure}[htbp]
\centering
\includegraphics[width=\textwidth]{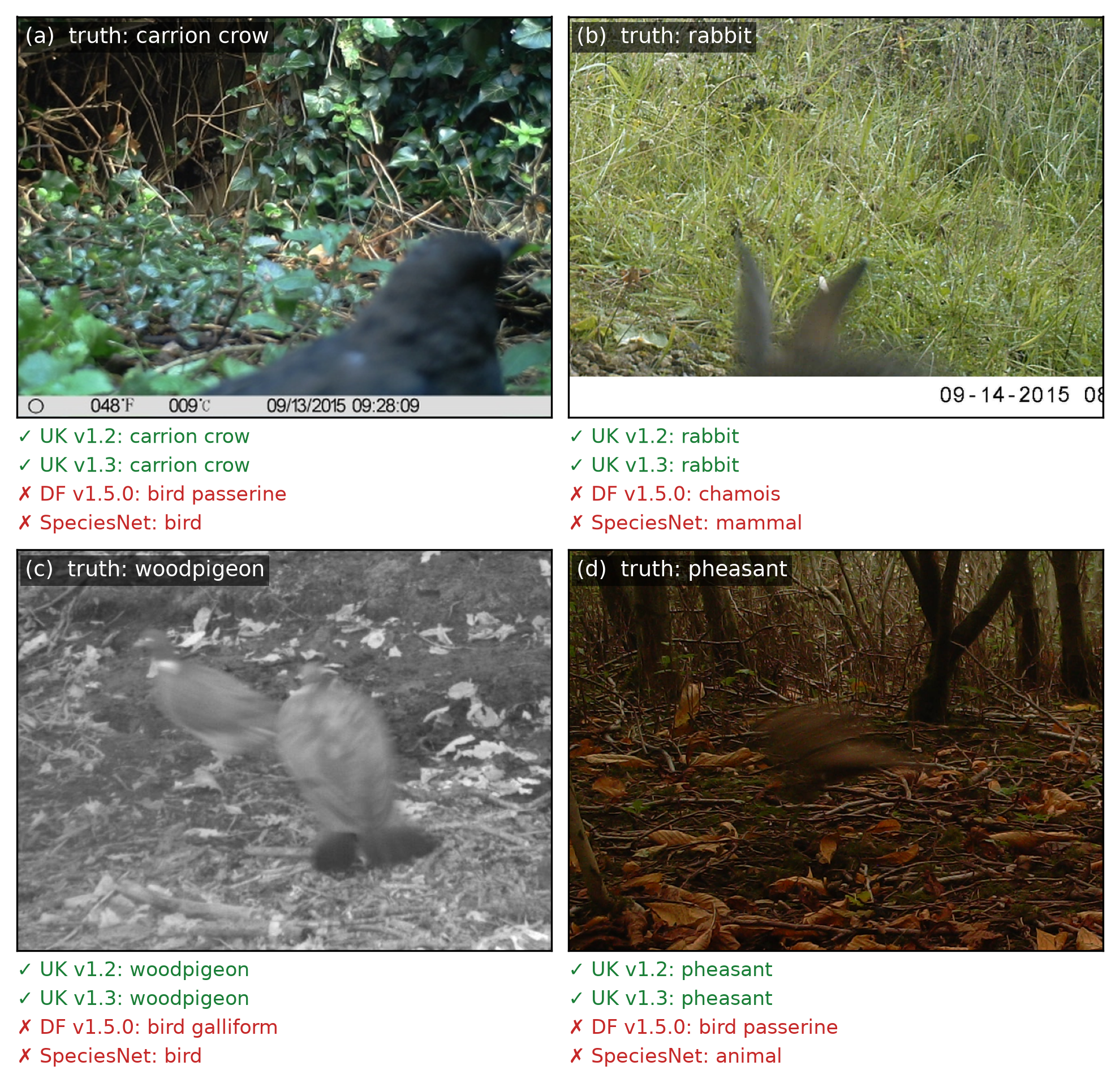}
\caption{Further images on which neither baseline returns the correct harmonised group while both DeepFaune-UK models name the exact species: (a) carrion crow, (b) rabbit, (c) woodpigeon and (d) pheasant. Symbols as in Figure A1.}
\label{fig:A3}
\end{figure}

Figure A4 shows errors removed by starting from the DeepFaune v1.5.0 (DINOv3) checkpoint: v1.2's confusion of roe with fallow deer (407 images, reduced to 126 in v1.3) and its tendency to label poorly lit or partial animals as cat or sheep. Figure A5 shows characteristic failures that remain. v1.3 greatly increased a confusion that was minor in v1.2, labelling 203 roe deer images as muntjac (42 for v1.2) (A5a). All three red squirrel images, which come from a single sequence, were labelled grey squirrel by both DeepFaune-UK models, the most consequential direction of error for the selective-trap application discussed in Section 5 (A5b). Some clearly visible animals were detected only below the detector's 0.25 confidence threshold and so were reported as empty (A5c), and detector false positives on empty frames were passed to the classifier and given an animal label (A5d), an error whose frequency the change of checkpoint did not affect (848 empty images labelled as animals by both versions).

\begin{figure}[htbp]
\centering
\includegraphics[width=\textwidth]{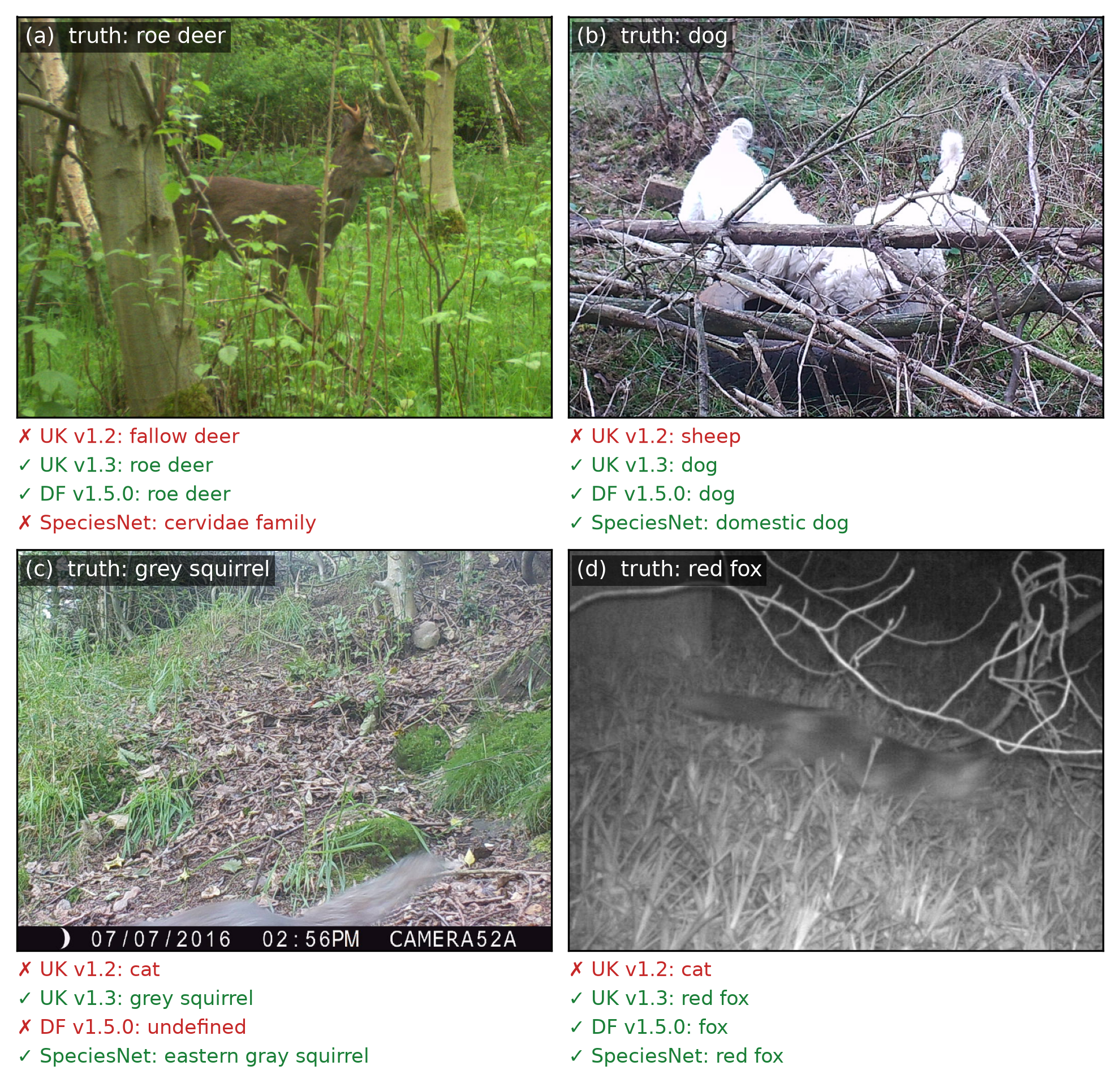}
\caption{Errors of DeepFaune-UK v1.2 corrected by v1.3: (a) roe deer labelled fallow deer, (b) dogs labelled sheep, (c) grey squirrel and (d) red fox labelled cat. Symbols as in Figure A1.}
\label{fig:A4}
\end{figure}

\begin{figure}[htbp]
\centering
\includegraphics[width=\textwidth]{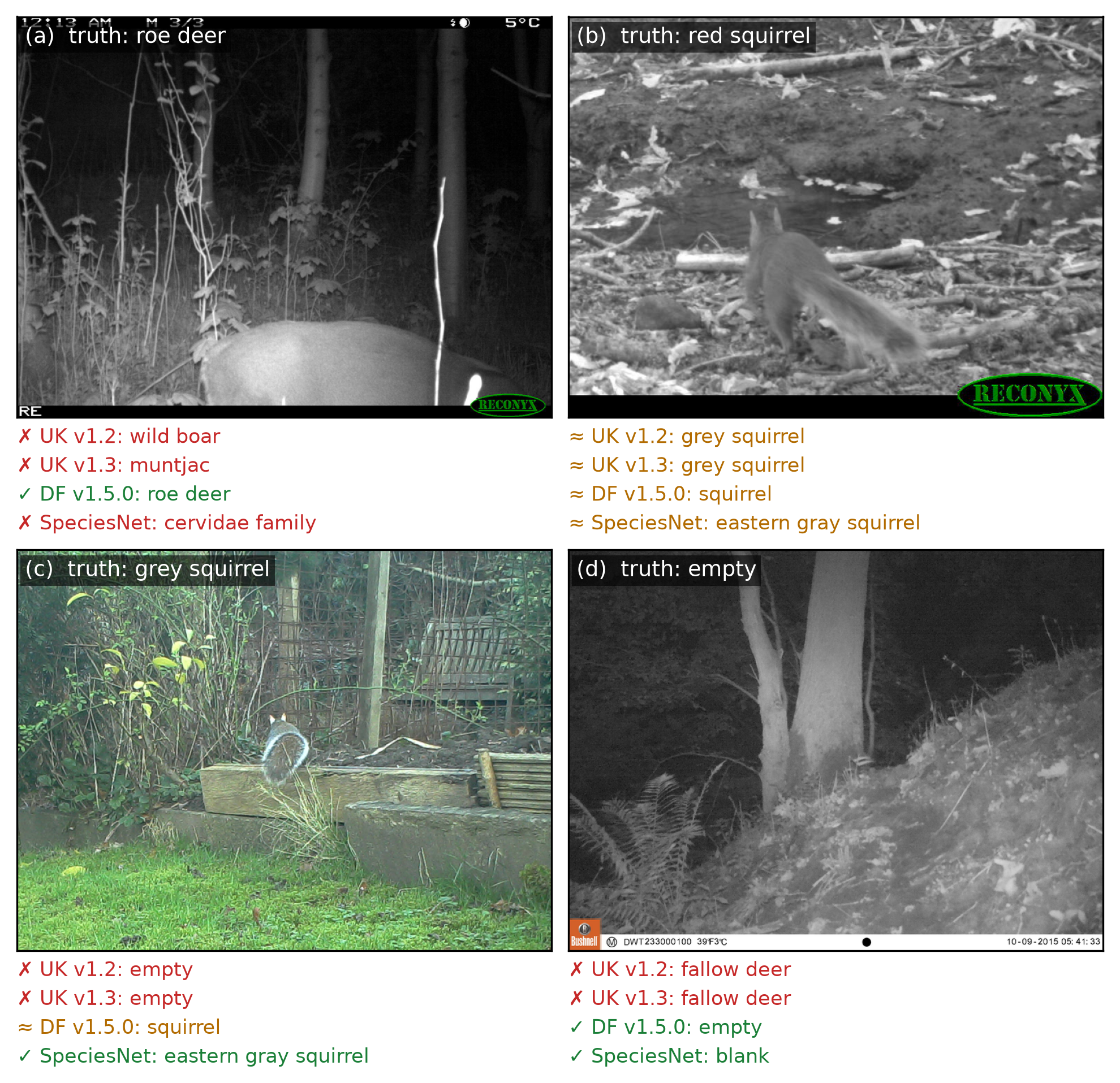}
\caption{Characteristic failures of DeepFaune-UK: (a) roe deer labelled muntjac by v1.3; (b) a red squirrel labelled grey squirrel by both versions; (c) a grey squirrel detected below the confidence threshold and reported as empty; and (d) an empty frame in which a false detection (on tree bark) was labelled fallow deer by both versions. Symbols as in Figure A1.}
\label{fig:A5}
\end{figure}

\section{Confidence Intervals and the Original Set}
\setcounter{figure}{0}
\renewcommand{\thefigure}{B\arabic{figure}}
\renewcommand{\theHfigure}{B\arabic{figure}}
\setcounter{table}{0}
\renewcommand{\thetable}{B\arabic{table}}
\renewcommand{\theHtable}{B\arabic{table}}

\begin{table}[htbp]
\centering
\caption{Point estimates and 95\% cluster-bootstrap confidence intervals (2,000 resamples) at the strict species level on the OOD evaluation set, resampling whole sequences or whole contributors (Section 3.6). Macro-F1 intervals are asymmetric because muntjac and red squirrel, each with three images from one sequence, are absent from 35--37\% of resamples and, when present, score an F1 of zero at this level for all models.}
\label{tab:B1}
\footnotesize\setlength{\tabcolsep}{4pt}
\begin{tabularx}{\textwidth}{>{\raggedright\arraybackslash}p{2.3cm}>{\centering\arraybackslash}X>{\centering\arraybackslash}X>{\centering\arraybackslash}X>{\centering\arraybackslash}X>{\centering\arraybackslash}X>{\centering\arraybackslash}X}
\toprule
\textbf{Model} & \textbf{Accuracy} & \textbf{95\% CI, sequences} & \textbf{95\% CI, contributors} & \textbf{Macro-F1} & \textbf{95\% CI, sequences} & \textbf{95\% CI, contributors} \\
\midrule
DeepFaune-UK v1.2 & 0.888 & 0.880--0.895 & 0.869--0.910 & 0.778 & 0.766--0.914 & 0.754--0.905 \\
DeepFaune-UK v1.3 & 0.906 & 0.899--0.913 & 0.889--0.924 & 0.786 & 0.775--0.923 & 0.768--0.919 \\
DeepFaune v1.5.0 & 0.565 & 0.550--0.580 & 0.399--0.715 & 0.398 & 0.395--0.466 & 0.390--0.466 \\
SpeciesNet 5.0.5 & 0.590 & 0.577--0.603 & 0.509--0.702 & 0.536 & 0.525--0.630 & 0.513--0.640 \\
\bottomrule
\end{tabularx}
\end{table}

\begin{table}[htbp]
\centering
\caption{Point estimates and 95\% cluster-bootstrap confidence intervals (2,000 resamples) at the like-for-like level on the OOD evaluation set, resampling whole sequences or whole contributors (Section 3.6). Calibrated rows use thresholds selected on the master validation split by the procedure of Section 4.2 (v1.2: detector 0.50, classifier 0.30; v1.3: 0.50 and 0.20, Section 4.3).}
\label{tab:B2}
\footnotesize\setlength{\tabcolsep}{4pt}
\begin{tabularx}{\textwidth}{>{\raggedright\arraybackslash}p{2.3cm}>{\centering\arraybackslash}X>{\centering\arraybackslash}X>{\centering\arraybackslash}X>{\centering\arraybackslash}X>{\centering\arraybackslash}X>{\centering\arraybackslash}X}
\toprule
\textbf{Model} & \textbf{Accuracy} & \textbf{95\% CI, sequences} & \textbf{95\% CI, contributors} & \textbf{Macro-F1} & \textbf{95\% CI, sequences} & \textbf{95\% CI, contributors} \\
\midrule
DeepFaune-UK v1.2 & 0.888 & 0.881--0.896 & 0.870--0.912 & 0.850 & 0.829--0.921 & 0.806--0.926 \\
DeepFaune-UK v1.3 & 0.906 & 0.899--0.913 & 0.890--0.925 & 0.857 & 0.837--0.929 & 0.817--0.934 \\
DeepFaune v1.5.0 & 0.914 & 0.909--0.920 & 0.902--0.922 & 0.855 & 0.829--0.930 & 0.818--0.931 \\
SpeciesNet 5.0.5 & 0.734 & 0.720--0.748 & 0.631--0.815 & 0.708 & 0.679--0.778 & 0.667--0.779 \\
DeepFaune-UK v1.2 (calibrated) & 0.889 & 0.881--0.896 & 0.869--0.911 & 0.837 & 0.825--0.915 & 0.812--0.913 \\
DeepFaune-UK v1.3 (calibrated) & 0.909 & 0.902--0.915 & 0.892--0.927 & 0.858 & 0.839--0.931 & 0.822--0.937 \\
\bottomrule
\end{tabularx}
\end{table}

\begin{table}[htbp]
\centering
\caption{Point estimates on the original 21,640-image set, before the removal of the 440 sequences that overlap with the master dataset (Section 3.5), for comparison with Table 1.}
\label{tab:B3}
\footnotesize\setlength{\tabcolsep}{4pt}
\begin{tabularx}{\textwidth}{>{\raggedright\arraybackslash}p{2.3cm}>{\centering\arraybackslash}X>{\centering\arraybackslash}X>{\centering\arraybackslash}X>{\centering\arraybackslash}X}
\toprule
\textbf{Model} & \textbf{Strict accuracy} & \textbf{Strict macro-F1} & \textbf{Like-for-like accuracy} & \textbf{Like-for-like macro-F1} \\
\midrule
DeepFaune-UK v1.2 & 0.894 & 0.782 & 0.895 & 0.854 \\
DeepFaune-UK v1.3 & 0.911 & 0.791 & 0.912 & 0.863 \\
DeepFaune v1.5.0 & 0.552 & 0.399 & 0.918 & 0.859 \\
SpeciesNet 5.0.5 & 0.605 & 0.548 & 0.747 & 0.719 \\
\bottomrule
\end{tabularx}
\end{table}
\end{document}